\documentclass{article} % For LaTeX2e
\usepackage{iclr2026_conference,times}

\usepackage{amsmath,amsfonts,bm}

\def\eqref#1{equation~\ref{#1}}
\def\1{\bm{1}}

\DeclareMathAlphabet{\mathsfit}{\encodingdefault}{\sfdefault}{m}{sl}
\SetMathAlphabet{\mathsfit}{bold}{\encodingdefault}{\sfdefault}{bx}{n}

\usepackage{hyperref}
\usepackage{url}
\usepackage{booktabs}       % professional-quality tables
\usepackage{amsfonts}       % blackboard math symbols
\usepackage{nicefrac}       % compact symbols for 1/2, etc.
\usepackage{xcolor}         % colors
\usepackage{graphicx}
\usepackage{multirow}
\usepackage[linesnumbered,ruled,vlined]{algorithm2e}

\usepackage{amsmath, amssymb}
\usepackage{etoolbox} % for \ifdefempty
\usepackage{makecell} % pretty stacked cells
\usepackage{arydshln}

\title{A Neurosymbolic Agent System for \\ Compositional Visual Reasoning}

\newcommand{\superscr}[1]{\texorpdfstring{$^{#1}$}{(#1)}} % typeset: ^{1}, PDF string: (1)
\author{Yichang Xu\superscr{1}, Gaowen Liu\superscr{2}, Ramana Rao Kompella\superscr{2}, Sihao Hu\superscr{1}, Fatih Ilhan\superscr{1},\\
\textbf{Selim Furkan Tekin}\superscr{1}, \textbf{Zachary Yahn}\superscr{1}, \textbf{Ling Liu}\superscr{1}\\
\textsuperscript{1}Georgia Institute of Technology, Atlanta, GA\\
\textsuperscript{2}Cisco Systems, USA \\
\{xuyichang,sihaohu,filhan,stekin6,zachary.yahn\}@gatech.edu \\
\{gaoliu,rkompell\}@cisco.com, ling.liu@cc.gatech.edu%
}

\iclrfinalcopy % Uncomment for camera-ready version, but NOT for submission.
\begin{document}

\maketitle

\vspace{-12pt}
\begin{abstract}
\noindent
\vspace{-2pt}
The advancement in large language models (LLMs) and large vision models has fueled the rapid progress in multi-modal vision-language reasoning capabilities. However, existing vision–language models (VLMs) remain challenged by compositional visual reasoning. This paper presents VLAgent, a neuro-symbolic approach to developing a Vision-Language Agent system for efficient compositional visual reasoning with three novel features. 
{\it First}, VLAgent develops an interpretable visualization-enhanced two-stage neuro-symbolic reasoning system. The first stage is managed by a front-end engine that generates a structured visual reasoning plan (symbolic program script) for each compositional visual reasoning task by utilizing a pre-trained LLM powered with few-shot chain-of-thought in-context learning. The second stage is managed by a high-performance back-end engine. It transforms the planning script into executable code based on visual input (image or video) and the combination of neural models and symbolic functions and then performs a sequence of actions for the compositional visual reason task. 
{\it Second}, to ensure and enhance the quality of mapping the logic plan to a sequence of executable instructions, VLAgent introduces the SS-parser, which examines the syntax and semantic correctness of the planning script, detects and repairs the logic errors found in the LLM-generated logic plan before generating the executable program. {\it Third}, VLAgent introduces the execution verifier in critical reasoning steps to validate and refine its compositional reasoning results in a stepwise manner, for example, ensemble methods for critical visual reasoning and caption analysis for low-confidence compositional reasoning. Extensive experiments were conducted on six visual benchmarks and compared to a dozen SoTA visual reasoning models. The results show that VLAgent outperforms existing representative approaches to compositional visual reasoning, while enabling self-interpretable visualization for human-in-the-loop debugging. Our code and runtime logs will be released upon acceptance.
\vspace{-6pt}
\end{abstract}

\vspace{-6pt}
\section{Introduction}
\label{sec:intro}
\vspace{-4pt}
Compositional visual reasoning tasks often involve a sequence of heterogeneous visual reasoning subtasks, and demand for multiple independently trained vision models or vision-language models (VLMs) to perform different subtask-specific visual reasoning. Furthermore, different visual reasoning tasks tend to require different compositions of multiple vision models in order to generate correct visual reasoning output. 
Hence, learning to perform diverse compositional visual reasoning tasks poses significant challenges to advanced large vision-language models, including GPT-4o, GPT-5. 
In this paper, we present VLAgent, a vision-language agent system that explores the neuro-symbolic approach to automatically breakdown each compositional visual reasoning task from end-users into a sequence of task-specific neuro-symbolic instructions in two stages. We argue that a neurosymbolic approach to compositional visual reasoning could be viewed as an attractive complementary representational learning framework to the advanced large vision-language models. Furthermore, we argue that to ensure high performance and high accuracy in reasoning output, a robust integration of neuro-symbolic learning for compositional reasoning requires the following four critical functional components to work in concert: plan generation, plan debugging and repair, verifiable plan execution with stepwise auditing, and output verification. By self-validation and self-refinement of neuro-symbolic instructions generated for compositional reasoning, prior to generating the final output, will notably fortify the generalization performance of VLAgent for complex visual reasoning.

\vspace{-2pt}
\noindent 
To date, research activities have been engaged towards complex visual reasoning along two interrelated threads. 
Neural Module Networks (NMN)~\cite{andreas2016neural,hu2018explainable,Hu_2017_ICCV,johnson2017inferring,andreas-etal-2016-learning} are pioneering in the model training category. This line of work aims to tackle the challenges of end-to-end visual reasoning models by decomposing complex reasoning tasks into modular compositional subroutines through supervised learning with large labeled training datasets. NMN development shows that neural modular networks can significantly improve the interpretability of visual reasoning. Inspired by the ideas of neural modular networks, recent approaches are centered on zero-shot learning, instead of training with supervised learning. Pre-trained LLMs (open source or close source) are utilized to generate structured programs for performing end to end compositional reasoning. For example, ProgPrompt~\cite{10161317} generates executable programs to help robots perform vision-related tasks.  ViperGPT~\cite{suris2023vipergpt} and VisProg~\cite{gupta2023visual} aim to solve visual question \& answer (VQA) problems with zero-shot learning. ViperGPT formulates Python programs based on existing Python libraries and VisProg uses LLM to generate program template embedding calls to external modules (pretrained models or preconfigured Python functions
%%and execute the LLM generated program through its Python engine 
, more detailed related work in Appendix~\ref{sec:related}). 
However,  existing visual reasoning methods suffer from a number of limitations. {\it First}, LLM generated programs often produce non-existent modules or logically flawed execution program steps. {\it Second}, existing methods lack of capability for checking the validity of LLM generated program w.r.t. both the feasibility of generating runnable code and the correctness of reasoning steps. As a result, existing methods tend to fail miserably when the LLM generated program is ill-formatted or logically incorrect due to unwanted hallucination. {\it Finally}, existing approaches often hard-wire a pre-defined external module for each reasoning step, making their performance bounded to the performance of the weakest external module(s) in the end-to-end reasoning process. 

Motivated by the above observations, we present VLAgent, a vision-language agent system for efficient end-to-end visual reasoning with three novel characteristics. {\it First}, the VLAgent develops a two-stage neuro-symbolic visual reasoning agent framework. Stage-1 managed by a front-end engine will utilize few-shot and Chain of Thought (CoT) in-context learning to finetune a pretrained LLM to learn to create a stepwise visual reasoning plan in the form of a structured logic program for any visual reasoning request from end-users. Stage-2 managed by a backend engine will support the mapping of each planning script to executable code and perform runtime execution to generate final reasoning output. 
{\it Second}, the VLAgent optimizes the two-stage neurosymbolic framework by developing the VLAgent SS-Parser, geared to improve the quality and correctness of LLM-generated planning scripts prior to generating executable code for runtime execution. This SS-Parser will inspect and repair syntactic and semantic errors detected in the planning program script. {\it Next}, the VLAgent enhances the generalization performance of the VLAgent by incorporating an output verifier to validate and refine compositional visual reasoning steps during runtime execution. 
%%, which further improves the generalization performance of the final output for each visual reasoning task. 
\begin{figure*} [htbp]
    \centering
    \vspace{-6pt}
    \includegraphics[width=0.8\linewidth]{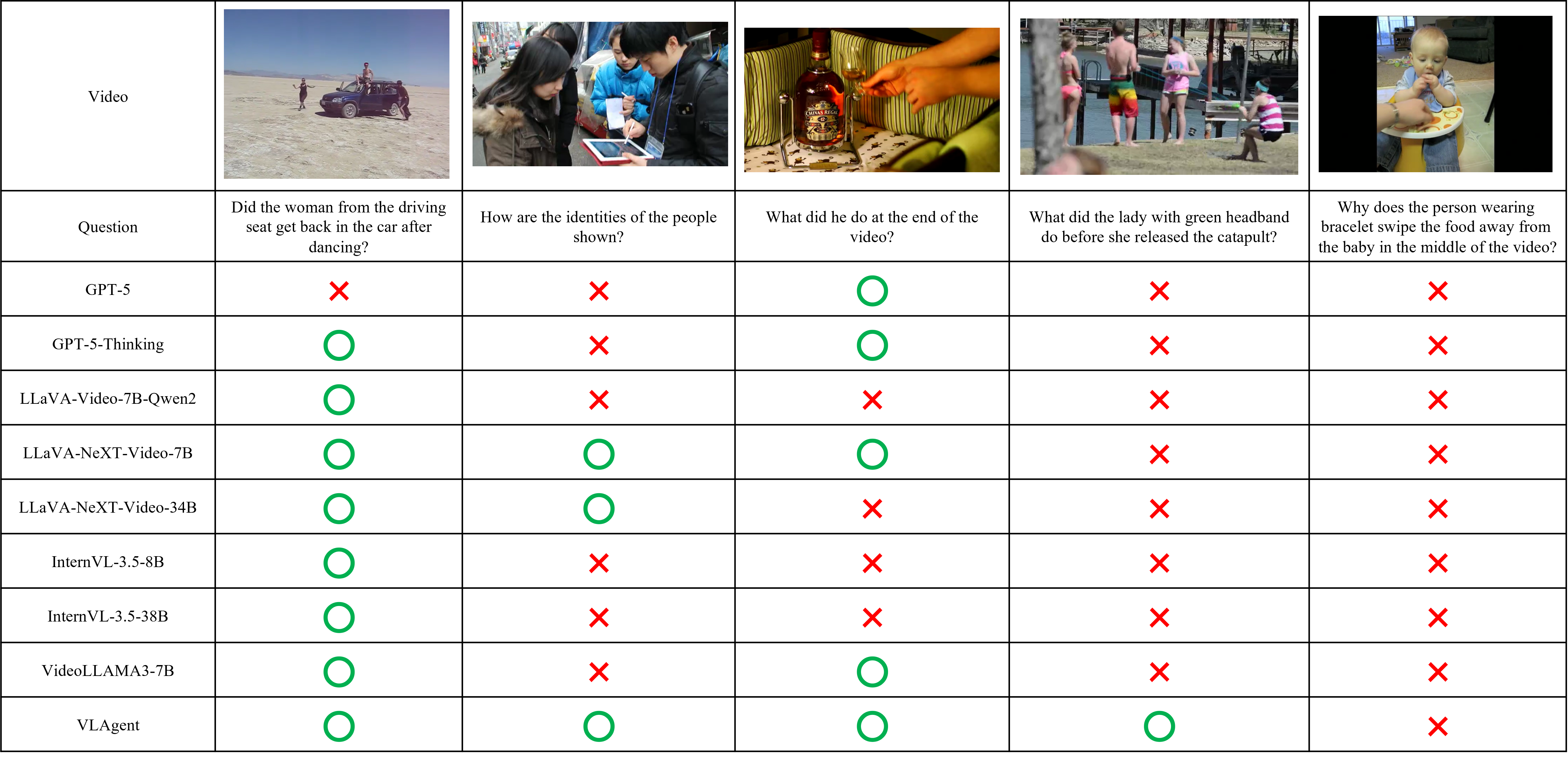}
       \vspace{-6pt}
    \caption{{\small \protect Examples illustrating performance of eight VLMs on NeXT-QA compared with VLAgent. 
    %%The raw video is named as ``\{video\_id\}.mp4'' and can be accessed through the \protect\href{https://drive.google.com/file/d/1jTcRCrVHS66ckOUfWRb-rXdzJ52XAWQH/view?pli=1}{NeXT-QA Google Drive}.
    }}
    \label{fig:five_cases_nextqa}
    \vspace{-4pt}
\end{figure*}
\\
\textbf{Figure~\ref{fig:five_cases_nextqa}} 
illustrates 
by examples the effectiveness of VLAgent compared with eight popular VLMs, including GPT-5 and GPT-5-Thinking on NeXT-QA benchmark~\cite{xiao2021next}. VLAgent correctly solves four out of five cases. 
%%demonstrates superior performance, 
Extensive experiments are performed on six 
representative visual reasoning benchmarks.
%%:
The results demonstrate that VLAgent consistently outperforms existing zero-shot learning methods by 3\%-40\% in terms of accuracy on image QA benchmarks and surpasses the representative zero-shot methods on video QA benchmarks by a margin of 7\%-19\%.
%\vspace{-10pt}
\section{Methodology}
\label{sec:method}
\vspace{-10pt}

\begin{figure*} [t!]
    \centering
    \vspace{-5pt}
    \includegraphics[width=\linewidth]{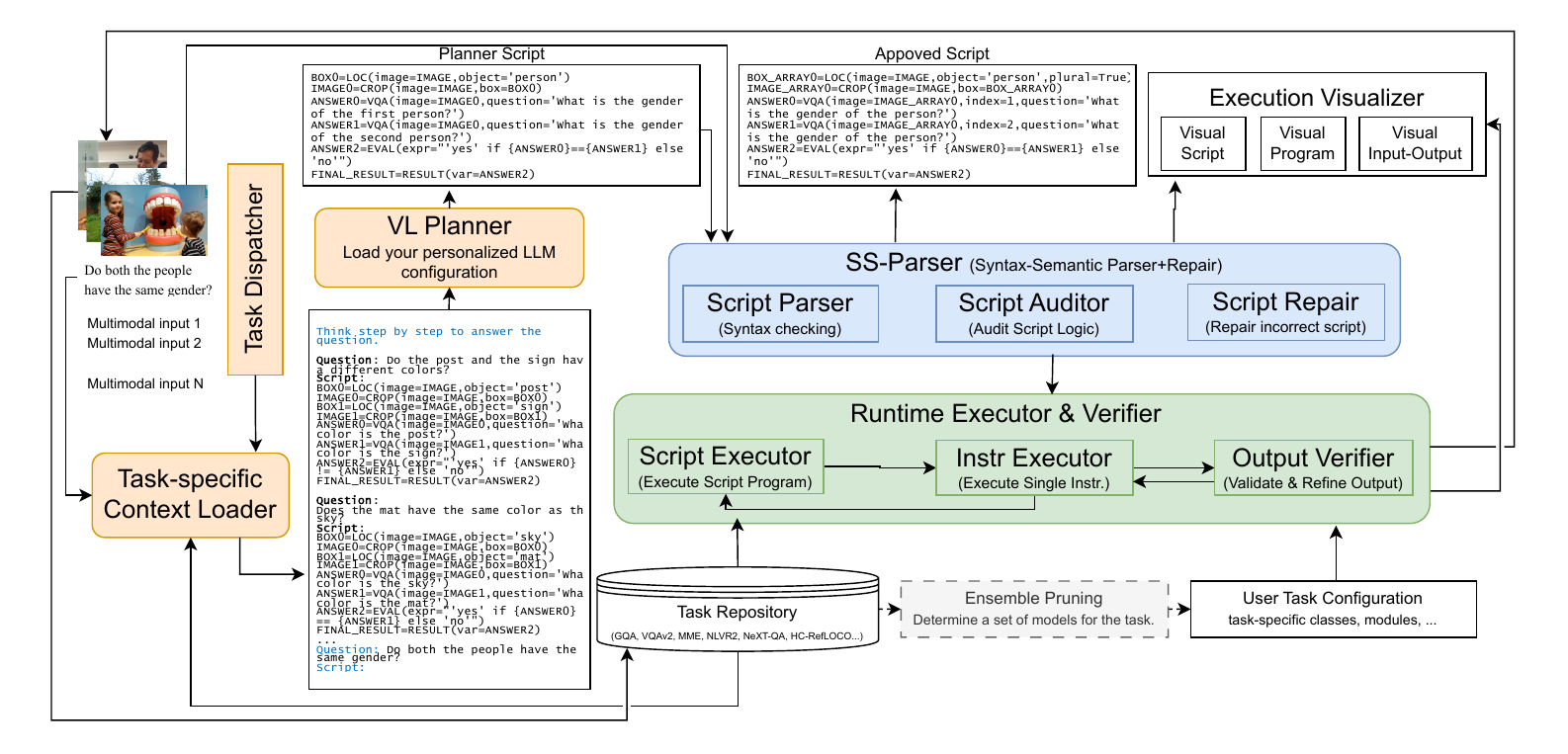}
    \vspace{-24pt}
    \caption{{\small A two-stage Neurosymbolic architecture of the VLAgent system with front-end engine and backend engine working in concert.}}
    %%. The output verifier may also exist in the path Step Executor to Scipt Executor when deep-learning-based models are invoked.}}
    \vspace{-14pt}
    \label{fig:architecture}
\end{figure*}

\begin{figure*} [h!]
    \centering
    \includegraphics[width=0.85\linewidth]{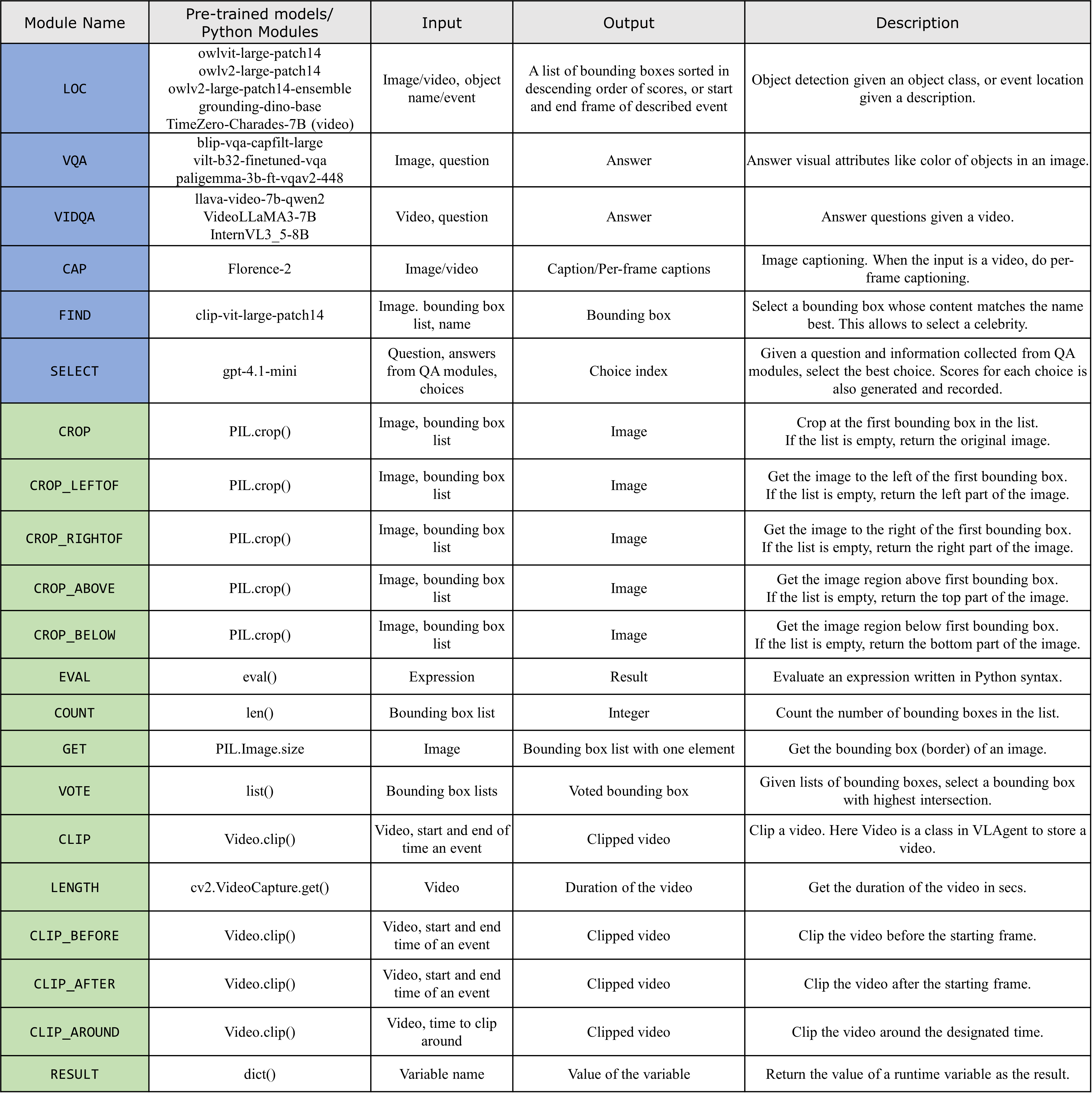}
    \vspace{-8pt}
    \caption{{\small Core Modules in VLAgent $\alpha$ release. 
   NLVR2 uses \texttt{VQA}, \texttt{EVAL} and \texttt{RESULT}. VideoQA uses modules taking video input plus \texttt{SELECT} and \texttt{EVAL}. HC-RefLOCO (referring expression) uses \texttt{LOC}, \texttt{CAP}, \texttt{FIND}, \texttt{VOTE}.}}
    \label{fig:modules}
    \vspace{-17pt}
\end{figure*}

\noindent 
We give an architectural overview of the VLAgent design methodology in \textbf{Figure~\ref{fig:architecture}}. The front-end engine of VLAgent is shown on the left, consisting of task dispatcher, task-specific context loader, and visual reasoning planner (aka script generator). When VLAgent receives a visual QA query task (e.g., ``{\em Do both the people have the same gender}''), the task-dispatcher will instruct the context-loader to compose the task-specific context with few-shot examples and CoT instructions as the prefix context for the visual reasoning planner to finetune a LLM with in-context learning (ICL). The LLM is either by default or chosen by the user of VLAgent in her configuration. 
Concretely, the planner will first prefix the visual QA query with the ICL context to compose a prompt query (see the middle rectangle under the planner by magnifying Figure~\ref{fig:architecture}) and then finetune the LLM via few-shot CoT in-context learning to generate a task-specific planning program script (an example in the top-middle rectangle). The planning script consists of a sequence of declarative program instructions, each is expressed as an assignment statement with output variable, per-instruction-specific visual reasoning module with module name and input parameters. 
All modules are either \textbf{external pre-trained models}, which can be by default or by user-choice in VLAgent configuration, or \textbf{internal modules} from the VLAgent Python library. 
\textbf{Figure~\ref{fig:modules}} shows a list of core modules supported in the alpha version of VLAgent and used in the experiments reported in this paper.   
%The module names are the naming convention learned by the Planner via a combo of $K$-shots and CoT for in-context instruction-learning. 
The first six rows are external pre-trained models marked in blue. The remaining modules marked in green, each corresponds to an internal module in the VLAgent Python library. \texttt{LOC}, \texttt{VQA}, \texttt{CAP}, \texttt{CROP} to \texttt{GET}, and \texttt{RESULT} are used for GQA, VQAv2, MME workloads.

The backend engine is the backbone of VLAgent. It takes both text input and visual input of each visual reasoning task, and examines the LLM-generated planning script through the SS-Parser (blue rectangle) and the Executor and Verifier (green rectangle).
The {\bf SS-Parser} (blue) consists of three main VLAgent modules: It first invokes the \textit{Script Parser} and \textit{Script Auditor} to examine the generated planning script and detect syntax errors and incorrect program logic. The \textit{Script Repair} module is triggered to fix errors and generate the correct planning script. Once the SS-Parser completes the syntax-semantic checking and repairing, it will forward the correctness approved script to the next VLAgent backbone subsystem: the {\bf Runtime Executor and Verifier}. It takes the SS-Parser-refined planning script with both text input and visual input and maps logic program script to the executable code, and then performs runtime execution with stepwise verification of reasoning correctness. It has three core modules. First, it invokes the \textit{Script Executor} to generate executable codes to call external modules or internal library functions. Next, it runs each executable instruction by the {\em Instruction Executor}. By running the executable instructions corresponding to the sequence of declarative program steps contained in the planning script, the executor will summarize and generate the final answer. To ensure the robustness and generalization performance of compositional reasoning of VLAgent, we validate the output from the executor through the VLAgent output verifier, which performs inconsistency resolution with statistical confidence scores. The result with the highest score is returned as the final answer that is returned to the user. 
The task repository and execution visualizer are shared by both front-end and backend engines of VLAgent. The task repository stores not only the ICL examples and CoT instructions, but also the backend neurosymbolic modules and configuration for different types of visual reasoning tasks. For example, users can add a new task by registering new modules and corresponding ICL examples and CoT instructions. The execution visualizer utilizes human-interpretable visual API for debugging and human-in-the-loop feedback analysis in both stages of the neurosymbolic reasoning process, configurable in VLAgent through the configuration manager.
With page limit, below we focus on 
%%two core components of VLAgent: 
the SS-Parser and the Plan Executor \& Verifier to describe the VLAgent design methodology.
%%and implementation detail. 
\vspace{-5pt}

\vspace{-3pt}
\subsection{\textbf{SS Parser}}
\vspace{-7pt}
We describe the design of three core components of our SS-Parser in this section with illustrative examples. First, the 
%%Before passing the planning script to the {\em Script Executor}, the VLAgent needs to perform syntax and semantic verification for detecting potential syntax errors and logic errors. The 
\emph{Script Parser} examines the LLM-generated planning script line by line to flag all syntax errors. The \emph{Script Auditor} performs the semantic-level examination to detect semantically incorrect parameters and semantically incorrect logic sequence of the planning steps. The {\em Script Repair} is designed to fix the errors and generate a logically consistent sequence of instructions. 
Typical syntax errors are the cases where a non-existent module name or false input parameters were used in the function call to external modules or internal modules. 
A frequent logic failure that many LLMs suffer is to make up some input parameter(s) in the \texttt{LOC} module: such as locating an object of ``standing'' or ``smiling'', and alike.   
We provide auto-scanner and auto-fixer to spot error type and error location in the planning script, and correct those errors accordingly, including type-checking of all external and internal modules w.r.t. module names, module input parameter types and output variable names, plus some error types for domain-dependent visual reasoning tasks, e.g., video vs image. 
%%Most of the syntax errors once detected can be corrected with high confidence. 
%%to see if the subject of the description is a noun  
%
\begin{figure*}[t!]
    \centering
     \vspace{-6pt}
    \includegraphics[width=0.7\linewidth]{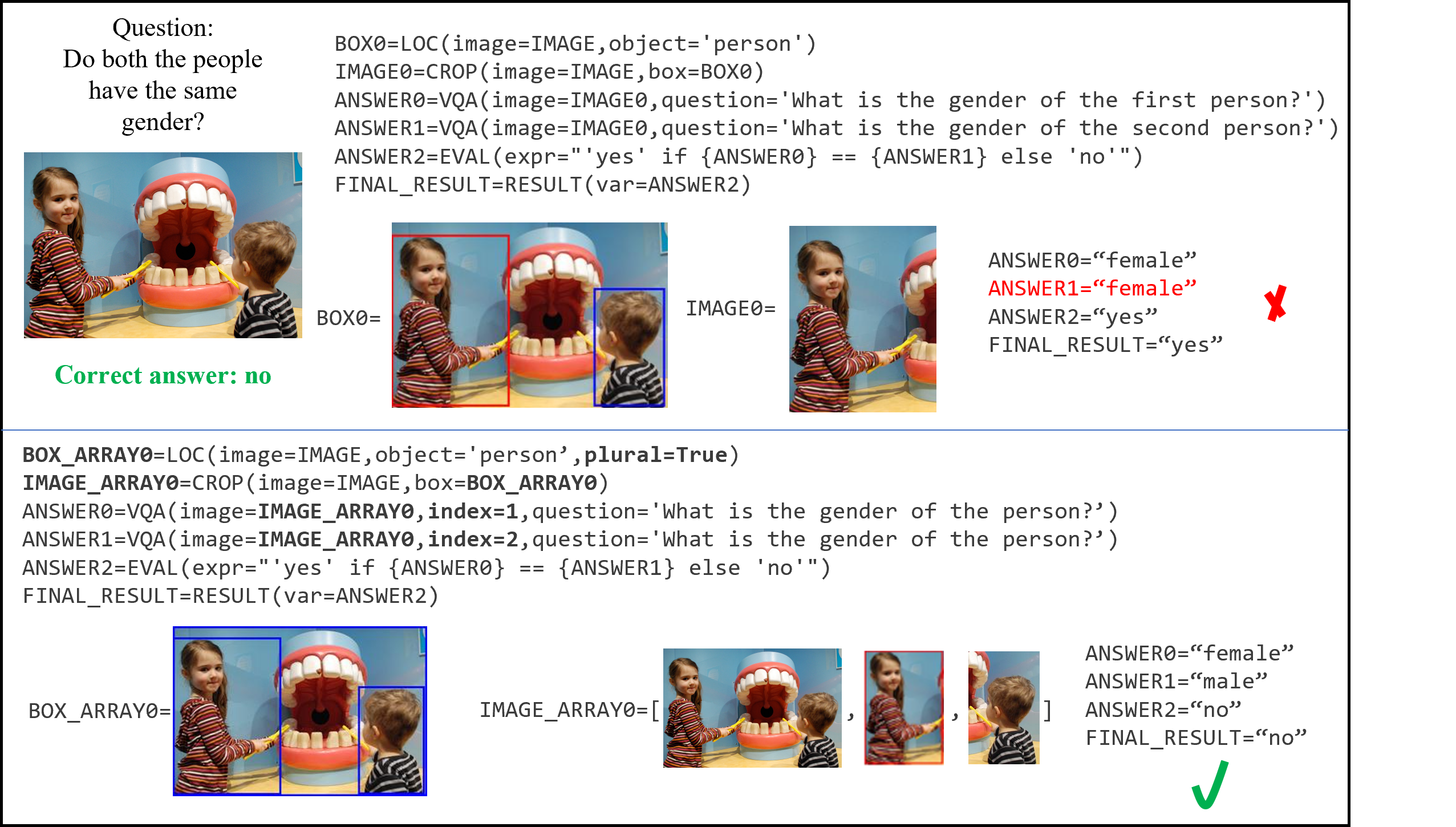}
    \vspace{-6pt}
    \caption{{\small VLAgent SS-Parser corrects the reasoning error detected in the LLM-generated planning script.}}
    \label{fig:parser}
    \vspace{-16pt}
\end{figure*}
%
%\vspace{-4pt}
\textbf{Figure~\ref{fig:parser}} illustrates by example the effect of SS-Parser on the overall performance of VLAgent.
Consider the example user query ``{\em Do both the people have the same gender}'' with the visual image (the top left). The ground truth answer for this zero-shot VQA is given under the image in green color. The planning script generated by LLM (GPT-3.5 in this case) is given at the top with six lines, calling \texttt{LOC}, \texttt{CROP}, \texttt{VQA}, \texttt{EVAL} and \texttt{RESULT} modules. The correct logic is to locate all the people in the image, crop them, ask the gender of each person and compare. 
Our SS-Parser spotted a logical error in the 4th line of the LLM-generated program, and fixed the error with the SS-parser refined script (shown in the bottom rectangle) prior to generating the executable. This results in correct final answer by VLAgent. In contrast, without using our SS-Parser, the execution of the LLM-generated script (in the top rectangle) resulted in a wrong answer. 
%%Later we will discuss the exact reason of the answer and our strategy to fix it. 
%%Concretely, we illustrate the utility of our SS-Parser in evaluating the correctness of LLM-generated planning script and repairing and correcting the planning script by the example in \textbf{Figure~\ref{fig:parser}}. 
Concretely, by examining the script at the top, although the intention of the script looks correct, there is a critical logical error: IMAGE0 is obtained by \texttt{CROP} of only the bounding box with the highest confidence score, which is the girl in the cropped image. Hence, the girl's gender is asked twice. 
%%A main source of the error is the inadequate use of the sequence of instructions when there are more than one objects of the same class to be detected (e.g., people in this query). 
Given two people are asked for comparison in this visual reasoning query, 
our SS-Parser fixed this problem by explicitly using an array as the output variable for \texttt{LOC} instruction and considering the bounding boxes with the top two highest scores, followed by performing \texttt{CROP} for each of the two bounding boxes in the output array of \texttt{LOC} respectively, it will capture two people. Followed by using \texttt{VQA} to ask the gender question on each of the two cropped images in \texttt{IMAGE\_ARRAY0}. This enables the \texttt{EVAL} module to compare the gender of the two people. The execution visualizer provides intuitive interpretation of the sequence of instructions performed for VLAgent to produce the correct final answer. 
Although we utilize this example to illustrate the SS-Parser's impact on VLAgent performance, the methodology of SS-Parser design, as we outlined earlier, involves a user-configurable set of pre-defined type-checking rules for both syntax and semantic errors w.r.t. external and internal module calls and the sequence of instruction steps with respect to the task-specific semantic context.

\vspace{-7pt}
\subsection{\textbf{\textbf Runtime Executor \& Output Verifier}}
\vspace{-6pt}
Recall the green rectangle in the right middle of \textbf{Figure~\ref{fig:architecture}}, this VLAgent subsystem consists of three core components.  The \emph{Script Executor} takes as the input an SS-parser approved planning script and maps it into executable instructions line by line, preserving the sequence of planning steps. The \emph{Instruction Executor} is called interatively to locate and run the corresponding external module or internal Python module.
%%(recall  \textbf{Figure~\ref{fig:modules}}).  
The output verifier performs stepwise evaluation of each visual reasoning instruction to determine whether and how ensemble methods are employed to verify and refine the per-instruction result, and caption-analysis is employed when the low confidence score is produced for an instruction-level visual-resoning subtask.  

\textbf{Caption Verifier}. In image QA or video QA tasks, for each image or video frame being  processed, if the result is in low confidence, we trigger the Caption-Verifier optimization to  
obtain the caption of the input image$/$video frame as a reference point for checking output consistency before generating our final answer.
%%for each user query. 
In the first prototype of VLAgent, we set Florence-2~\cite{Xiao_2024_CVPR} as the default external module to implement our caption verifier, reconfigurable in VLAgent initialization. The caption-verifier first generates a detailed caption of the image or the given video frame, which includes the key elements and their states. Next, it utilizes a pretrained LLM like GPT-3.5 (chat model) to assess whether the caption provides sufficient clues to infer the correct answer. If the caption contains the necessary information, the comparison of the caption-inferred answer is made with the script executor's output. If the comparison results in a match, indicating the consistency between the execution output and the caption-inferred answer, the final answer is regarded as highly reliable. Otherwise, VLAgent will analyze explicit details in the caption and the caption-derived answer is chosen as the final result if explicit and accurate clues are found in the caption analysis. Our ablation study in Experiments section 
%(T\textbf{able~\ref{tab:GQA_ablation}}) -- Yichang this table generates latex error
shows the improvement brought by the caption verification.
%%overall brings non-negligible benefit 
%%(see Table~\ref{tab:GQA_ablation}) Yichang, error in latex for this table.
We also provide a video QA example 
(Figure~\ref{fig:video_example} of Appendix~\ref{sec:additional_eg}) 
to show the detailed procedure of how our caption-verifier corrects a wrong result produced by our script executor.

\textbf{Ensemble Verifier}. VLAgent improves external-module reliability (e.g., \texttt{LOC}, \texttt{VQA}) by leveraging the fusion of multiple independently pre-trained models but develops ensemble selection algorithm to prune a pool of $N$ candidate models to a small subset of \(M\) models for efficiency ($M<<N$)~\cite{tekin-etal-2024-llm}.
%%%%%%%%%%%%%%%%%%%%%%%%%%%%%
Let \(P\) number of model calls to evaluate an ensemble, e.g., \(P=100\), 
each model \(j\) returns \(B_{ij}\) for \texttt{LOC}. We fuse all candidates per call to obtain \(B_i\). The agreement and score are
\[
I_{ij}=\frac{|Area(\cup B_{ij})\cap Area(\cup B_i)|}{|Area(\cup B_{ij})\cup Area(\cup B_i)|},
\qquad
s_j=\frac{1}{P}\sum_{i=1}^{P} I_{ij}.
\]
For \texttt{VQA}, set \(I_{ij}=\mathbb{I}[A_{ij}\cap A_i\neq\emptyset]\); for \texttt{VIDQA}, set \(I_{ij}=\mathbb{I}[choice_{ij}=choice_i]\). We cluster \(\{s_j\}_{j=1}^{N}\) with K-Means \cite{macqueen1967multivariate}, choose the number of clusters via the Silhouette criterion \cite{rousseeuw1987silhouettes}, and iteratively add the cluster(s) with the largest scores until reaching \(M\) models for inference \cite{chen2025lvagent}. A detailed algorithm and notation is %%explanation of notations and the algorithm is illustrated 
in Appendix~\ref{sec:pruning}.

\textbf{Long Video Optimization.\/} Another critical optimization is resource-aware development of efficient neurosymbolic reasoning agent. Consider video QA tasks, we first develop video sampling methods through video partitioning into several chunks and then identifying important chunks via sampling. When using \texttt{LOC} module to locate an event in time dimension and use \texttt{VIDQA} module to perform  video question answering, VLAgent feeds into these modules only a sequence of sampled frames rather than the entire video as input. For long videos, 
%% Due to the complexity of these models, 
the sampling rate is usually much lower than the video frame rate. Therefore, 
%%when it comes to long videos, we need to do some optimization 
we employ some optimizations to reduce or minimize the information loss from a low sampling rate. 
%%We have a configuration to decide whether the video is long. 
Suppose that we sample at most $f$ frames in these models, and we wish the sampling rate is at least $v$ fps. For a video of length $l$ (seconds), we uniformly divide it into $\lceil vl/f\rceil$ chunks. Note that $\lceil vl/f\rceil>1$ in most scenarios under low sampling rate $f$, say 4 or 6 frames per video or per video chunk, we perform the following preprocessing: For each video, we sample $f$ frames uniformly and use a video QA model to generate the caption of the video. Then according to the question, an ensemble of multiple LLMs (e.g., GPT-4.1-mini) is used to judge whether only part of the video needs to be watched and leverages the sampled frame inference results to produce statistical scores on which chunks are more relevant to answering the question. We resample another $f$ frames from the video or from each chunk to perform the same procedure when the ensemble result is undesiable, e.g., no chunk is highly relevant or the entire video should be watched. We constrain this iterative procedure by at most $t$ iterations (e.g., $t=3$). 
%%Once we concluded that only 
Given a small number of selected chunks 
%of the long video 
needs to be watched for the visual reasoning question, we concatenate the selected chunks following their inherent temporal sequence and send it to the visual reasoning planner to follow the neurosymbolic procedure of VLAgent for compositional reasoning. Figure~\ref{fig:video_example} in Appendix~\ref{sec:additional_eg} illustrates this optimization with an example.

\vspace{-7pt}
\section{Experiments}
\label{sec:exp}
\vspace{-7pt}
We report the evaluation of VLAgent and the comparison with the SOTA representative visual reasoning methods on six popular benchmarks: 
%%four in the image QA domain and two in video QA domain. They are 
\texttt{GQA}~\cite{hudson2019gqa}, \texttt{NLVR2}~\cite{suhr2018corpus}, \texttt{VQAv2}~\cite{goyal2017making}, \texttt{MME}~\cite{fu2024mmecomprehensiveevaluationbenchmark} (existence, position, color category),  \texttt{NeXT-QA}~\cite{xiao2021next} and 
%%video QA benchmark and referring expression benchmark 
\texttt{HC-RefLOCO}~\cite{10.5555/3737916.3740138}.
All experiments are conducted with H100$/$H200 GPU in a Python 3.9 environment.
\vspace{-8pt}
\subsection{Experimental Comparison Results}
\vspace{-8pt}
\noindent 
%%This section reports on the experimental comparison of VLAgent with representative SOTA approaches on 6 benchmarks. 
{\bf Table~\ref{tab:comparison}} compares VLAgent with representative image QA approaches in VLMs category and zero-shot methods on 4 popular ImageQA benchmarks. 
%%We use the reported accuracy in the respective paper for fair comparison. 
We observe that existing zero-shot methods, represented by ViperGPT~\cite{suris2023vipergpt} and Visprog~\cite{gupta2023visual}, exhibit a significant performance gap compared to supervised fine-tuning approaches on all 4 benchmarks. In comparison, VLAgent (zero-shot) achieves superior performance for NLVR2 and MME by $0.3\sim29.7$\% and $1.0\sim30.3$\% respectively,  comparable performance for GQA and VQAv2. 
For VQAv2, VLAgent outperforms the three SOTA zero-shot methods (ViperGPT, VisProg and GENOME) and 10 VLM methods by $1.6\sim39.5$\%. Compare against the supervised model BLIP-VQA-Capfilt~\cite{li2022blip} (directly finetuned on VQAv2), VLAgent achieves 76.9\% accuracy compared to the best VLM of 75.3\%, reducing the gap to about 4\%. 
%%and outperforms the other VLMs by at most $30$\%. 
%on the NLVR2 leaderboard~\cite{nlvr2homepage}. 
For GQA, VLAgent achieves an accuracy of 61.9\%, which outperforms all 3 zero-shot methods by $7.6\sim26.4$\%, and outperforms BLIP-VQA-Capfilt (supervised) and 4 other VLMs 
%%Vilt-b32 and PaliGemma2 
by $3.8\sim 17.4$\%. VLAgent offers on-par performance to InternVL3, and narrows the gap of all zero-shot methods to LLaVA-1.5-7B (70.3\%), MiniCPM-V-4.5 (70.8\%), Kimi-VL-A3B (71.1\%) and Phi-3.5 (72.7\%) by a large margin (61.9\% vs $35.5\sim 54.3$\%). 
\vspace{-10pt}
{\scriptsize 
\begin{table}[h!]
 \centering
 \vspace{-5pt}
 \caption{{\small Comparing VLAgent with 14 representative methods (VLMs and Zero-Shot) on 4 benchmarks. Improvement min and max are the lower and upper bound on improvement over 3 ZS and 10 VLMs methods.}}
 {\small
 \begin{tabular}{c|c|c|c|c}
 \toprule
 Method & NLVR2 & GQA & VQAv2 & MME \\
 \midrule
 BLIP-VQA-Capfilt~\cite{li2022blip} (sup) & 44.8\% & 54.5\% & {\bf 81.2\%} & 82.8\%\\
 Vilt-b32~\cite{kim2021vilt} (VLM) & 46.6\% & 51.4\% & 73.6\% & 74.8\%\\
 PaliGemma2~\cite{steiner2024paligemma} (VLM) & 44.4\% & 44.5\% & 46.5\% & 61.6\% \\
Gemma3-12B~\cite{sellergren2025medgemma} (VLM) & {\bf 73.8\%} & 60.5\% & 66.5\% & 82.4\% \\
SmolVLM~\cite{marafioti2025smolvlm} (VLM) & 51.0\% & 58.1\% & 71.0\% & 84.7\% \\
 InternVL3~\cite{chen2024internvl} (VLM) & 50.6\% & 61.9\% & 61.4\% & 84.1\%\\
 Idefics2~\cite{laurençon2024matters} (VLM) & 50.2\% & 63.2\% & 74.0\% & 86.6\%\\
 LLaVA-1.5-7B~\cite{liu2023improvedllava} (VLM) & 50.3\% & 70.3\% & 71.0\% & 85.0\%\\
 MiniCPM-V-4.5~\cite{yao2024minicpm} (VLM) & 51.2\% & 70.8\% & 68.9\% & 85.7\% \\
Kimi-VL-A3B~\cite{kimiteam2025kimivltechnicalreport} (VLM) & 51.0\% & 71.1\% & {\bf 75.3\%} & {\bf 87.4\%} \\
 Phi-3.5~\cite{abdin2024phi} (VLM) & 64.5\% & {\bf 72.7\%} & 65.5\% & 86.0\% \\
 \midrule
 ViperGPT~\cite{suris2023vipergpt} (ZS) & $-$ & 35.5\% & 37.4\% & 58.1\%\\
 VisProg~\cite{gupta2023visual} (ZS) & 69.3\% & 54.3\% & 72.3\% & 85.3\%\\
 GENOME~\cite{chen2023genome} (ZS) & $-$ & 44.7\% & 60.5\% & 77.9\% \\
 \midrule
 VLAgent (ZS) & {\bf 74.1\%} & {\bf 61.9\%} & {\bf 76.9\%} & {\bf 88.4\%} \\
 \hdashline
 {\bf Improvement on ZS (min)} 
   & {\bf \textcolor{green!70!black}{+4.8\%}} & {\bf \textcolor{green!70!black}{+7.6\%}} & {\bf \textcolor{green!70!black}{+4.6\%}} & {\bf \textcolor{green!70!black}{+3.1\%}} \\
   {\bf Improvement on ZS (max)} & {\bf \textcolor{green!70!black}{+4.8\%}} & {\bf \textcolor{green!70!black}{+26.4\%}} & {\bf \textcolor{green!70!black}{+39.5\%}} & {\bf \textcolor{green!70!black}{+30.3\%}} \\
 \hdashline
{\bf Improvement (min)} 
   & {\bf \textcolor{green!70!black}{+0.3\%}} & - & {\bf \textcolor{green!70!black}{+1.6\%}} & {\bf \textcolor{green!70!black}{+1.0\%}} \\
   {\bf Improvement (max)} & {\bf \textcolor{green!70!black}{+29.7\%}} & {\bf \textcolor{green!70!black}{+26.4\%}} & {\bf \textcolor{green!70!black}{+39.5\%}} & {\bf \textcolor{green!70!black}{+30.3\%}} \\
 \bottomrule
 \end{tabular}%
 }
 \label{tab:comparison}%
 \vspace{-10pt}
\end{table}%
}

We next evaluate the generalization capability of VLAgent on two recent video benchmarks: NeXT-QA~\cite{xiao2021next} and HC-RefLOCO~\cite{10.5555/3737916.3740138}. For NeXT-QA video understanding benchmark, we sample 200 questions per-type from the test set, resulting in a test set of 1493 questions. {\bf Table~\ref{tab:nextqa_results}} shows the comparison result of VLAgent with six VLMs and five VideoQA specific agent-based methods (zero-shot). VLAgent exhibits a high accuracy on all hard splits and achieves an overall accuracy of 76.0\%, surpassing all 11 methods compared with the gain margin of $2.0 \sim 19.1$\%.
\begin{table}[h!]
  \centering
  \caption{{\small Performance of VLAgent compared with 11 representative methods in VLMs category (top 6 rows) and zero-shot agent methods (middle 5 rows). Hard Split-T means temporal reasoning questions and Hard Split-C means causal reasoning questions. "Overall" represents for the overall accuracy of all types of questions.}}
  {\small
    \begin{tabular}{c|c|c|c}
    \toprule
         Method & Hard Split-T & Hard Split-C & Overall \\
    \midrule
     LLaVA-Video-7B-Qwen2~\cite{zhang2024videoinstructiontuningsynthetic} & 60.4\%  & {\bf 69.5\%}  & 70.9\% \\
    LLaVA-NeXT-Video-7B~\cite{zhang2024llavanextvideo}  & 54.8\%  & 65.3\%  & 65.8\% \\
    LlaVA-NeXT-Video-34B~\cite{zhang2024llavanextvideo} & 55.0\%    & 63.8\%  & 61.8\% \\
    InternVL-3.5-8B~\cite{wang2025internvl3_5}  & 58.6\%  & 66.5\%  & 68.1\% \\
    InternVL-3.5-38B~\cite{wang2025internvl3_5} & 61.1\%  & 65.8\%  & 68.8\% \\
    VideoLLaMA3-7B~\cite{damonlpsg2025videollama3} & \textbf{66.3\%} & 68.0\%    & {\bf 72.2\%} \\
    \midrule
    ViperGPT~\cite{suris2023vipergpt} (ZS) & 48.7\%  & 56.2\%  & 56.9\% \\
    SeViLA~\cite{yu2023self} (ZS) & 59.6\%  & 58.5\%  & 64.0\% \\
    VideoAgent~\cite{VideoAgent} (ZS) & 60.0\%    & 68.3\%  & 66.1 \%\\
    TravelLER~\cite{shang2024traveler} (ZS) & 56.9\%  & 65.4\%  & 66.0\% \\
    MoReVQA~\cite{min2024morevqa} (ZS) & 64.6\%  & 70.2\%  & 69.2\% \\
    \midrule
        VLAgent & \textbf{67.3\%}  & \textbf{74.8\%} & \textbf{76.0\%} \\
    \hdashline
    {\bf Improvement on ZS (min)}
      & {\bf \textcolor{green!70!black}{+2.7\%}} & {\bf \textcolor{green!70!black}{+4.6\%}} & {\bf \textcolor{green!70!black}{+6.8\%}} \\
      {\bf Improvement on ZS (max)} & {\bf \textcolor{green!70!black}{+18.6\%}} & {\bf \textcolor{green!70!black}{+18.6\%}} & {\bf \textcolor{green!70!black}{+19.1\%}} \\
    \hdashline
    {\bf Improvement (min)}
      & {\bf \textcolor{green!70!black}{+1.0\%}} & {\bf \textcolor{green!70!black}{+4.6\%}} & {\bf \textcolor{green!70!black}{+2.0\%}} \\
      {\bf Improvement (max)} & {\bf \textcolor{green!70!black}{+18.6\%}} & {\bf \textcolor{green!70!black}{+18.6\%}} & {\bf \textcolor{green!70!black}{+19.1\%}} \\
    \bottomrule
    \end{tabular}%
    }
  \label{tab:nextqa_results}%
\end{table}%
For HC-RefLOCO~\cite{10.5555/3737916.3740138}, representing the complex referring expression benchmark on long descriptions, we sampled 1400 referring expressions from the test set to test VLAgent. Unlike traditional referring expression datasets like RefCOCO~\cite{kazemzadeh-etal-2014-referitgame}, which are already saturated in terms of performance due to the emergence of expression-based grounding models (e.g., OWLV2~\cite{minderer2023scaling} and Grounding Dino~\cite{liu2023grounding} used in this paper), HC-RefLOCO grounds a person with fine-grained description of appearance, position, movement, and so forth, posing a greater challenge to grounding models.
%challange 
For each grounding task in HC-RefLOCO, we perform the caption-verification as follows: we first use the LLM to identify the person whose caption best matches the description; if no match is found, then we skip the verification, otherwise check whether the bounding box of this person substantially overlaps with the predicted result. If yes, the prediction is considered correct. Otherwise, we compare the caption of the predicted bounding box with the description to make the reasoning for the final alignment. We correct the bounding box only when it aligns significantly worse than the caption of the originally selected person.
{\bf Table~\ref{tab:refloco_result}} shows the performance of VLAgent on HC-RefLOCO compared to the 13 representative methods, of which SPHINX-v2-1K~\cite{lin2023sphinx} is the well-known SOTA method. Following the metrics in HC-RefLOCO, 
%%~\cite{10.5555/3737916.3740138}, 
Acc0.5, Acc0.75, Acc0.9 and mAcc are used to measure and compare the performance. AccX means the ratio of test cases where the IoU between the predicted box and the ground truth is greater than X, and mAcc is the average value of Acc0.5 through Acc0.95 with a step size of 0.05. We observe that VLAgent surpass the 13 SOTA approaches compared in Acc0.75, Acc0.9 and mAcc, while using lightweight object detection models for inference time efficiency, and 
%%Especially, 
%VLAgent has 
a performance gain over SPHINX-v2 (the best at Acc0.9) by about 7\%. 
\vspace{-12pt}
\begin{table}[h!]
  \centering
  \caption{{\small Comparison of VLAgent on HC-RefLOCO with 13 SOTA methods (zero-shot or VLMs).}}
  {\small
    \begin{tabular}{c|c|c|c|c}
    \toprule
    Model & Acc0.5 & Acc0.75 & Acc0.9 & mAcc \\
    \midrule
    GPT-4V~\cite{openai2023gpt4} & 17.4\%  & 2.6\%   & 0.3\%   & 5.5\% \\
    GroundingGPT~\cite{li-etal-2024-groundinggpt} & 56.6\%  & 27.2\%  & 5.3\%   & 29.8\% \\
    Ferret 13B~\cite{you2023ferret} & 52.9\%  & 38.5\%  & 15.6\%  & 35.7\% \\
    KOSMOS-2~\cite{peng2024grounding} & 45.3\%  & 38.0\%    & 20.0\%    & 34.1\% \\
    Qwen-VL~\cite{Qwen-VL} & 67.9\%  & 56.8\%  & 34.8\%  & 52.8\% \\
    OFA-Large~\cite{wang2022ofa} & 70.5\%  & 61.6\%  & 44.0\%    & 58.1\% \\
    SPHINX~\cite{lin2023sphinx} & 77.5\%  & 61.0\%    & 27.0\%    & 55.4\% \\
    SPHINX-1K~\cite{lin2023sphinx} & 80.7\%  & 68.6\%  & 41.1\%  & 63.0\% \\
    SPHINX-v2-1K~\cite{lin2023sphinx} & \textbf{84.1}\%  & 77.1\%  & 56.2\%  & 71.7\% \\
    PixelLM 13B~\cite{ren2024pixellm} & 63.6\%  & 46.6\%  & 25.8\%  & 44.6\% \\
    LISA~\cite{lai2024lisa}  & 52.4\%  & 42.1\%  & 31.3\%  & 41.1\% \\
    PSALM~\cite{zhang2024psalm} & 61.7\%  & 53.6\%  & 40.2\%  & 51.1\% \\
    GlaMM~\cite{rasheed2024glamm} & 66.1\%  & 56.9\%  & 44.2\%  & 55.0\% \\
    \midrule
        VLAgent & 82.6\%  & \textbf{77.4}\%  & \textbf{63.2}\%  & \textbf{73.9}\% \\
    \hdashline
    {\bf Improvement (min)}
      & - & {\bf \textcolor{green!70!black}{+0.3\%}} & {\bf \textcolor{green!70!black}{+7.0\%}} & {\bf \textcolor{green!70!black}{+2.2\%}} \\
      {\bf Improvement (max)} & {\bf \textcolor{green!70!black}{+65.2\%}} & {\bf \textcolor{green!70!black}{+74.8\%}} & {\bf \textcolor{green!70!black}{+62.9\%}} & {\bf \textcolor{green!70!black}{+68.4\%}} \\
    \bottomrule
    \end{tabular}%
    }
  \label{tab:refloco_result}%
\end{table}%

\vspace{-18pt}
\subsection{Ablation Study}
\label{sec:ablation}
% \vspace{-6pt}
We compare the naive VLAgent (without SS-Parser+Caption-verifier+Ensemble verifier) with VLAgent+SS-Parser+Caption-verifier, and the full fledged VLAgent on four popular LLMs: gpt-3.5-turbo-instruct~\cite{openai2023gpt35turboinstruct}, Mistral-Small-24B-Base-2501~\cite{jiang2023mistral7b}, GLM4-9B~\cite{glm2024chatglm}, and Llama3-8B~\cite{grattafiori2024llama3herdmodels}. {\bf Table~\ref{tab:GQA_ablation}} reports the results on GQA. The performance of VLAgent improves progressively with the addition of SS-Parser and caption-verifier (row 2) and the addition of ensemble-verifier (row 3), compared to the naive version of VLAgent without SS-parser and output verifiers. The performance gains of VLAgent powered by our SS-Parser and Output Verifiers are consistent across the 4 popular LLMs for generating the planning scripts. Similar observations are made on other benchmarks as well (see Appendix~\ref{sec:ablation_llm}). An ablation study on inference latency is provided in Appendix~\ref{sec:latency}.
\vspace{-18pt}
%%thereby illustrating the effectiveness of our approach. 
%
% Table generated by Excel2LaTeX from sheet 'Sheet2'
\newcommand{\gainnaive}[1]{\textbf{\textcolor{green!60!black}{#1}}}
\begin{table}[h!]
  \centering
  \caption{{\small Ablation study (GQA)}}
  {\small
  \begin{tabular}{c|c|c|c|c}
    \toprule
    Method & GPT 3.5 & Llama & Mistral & GLM \\
    \midrule
    VLAgent naive & 54.4\% & 54.1\% & 55.2\% & 54.7\% \\
    \hdashline
    VLAgent +parser+cap-verf
      & \makecell{58.9\%\\\gainnaive{+4.5\%}}  % vs naive
      & \makecell{56.9\%\\\gainnaive{+2.8\%}}
      & \makecell{58.6\%\\\gainnaive{+3.4\%}}
      & \makecell{58.2\%\\\gainnaive{+3.5\%}} \\
    \midrule
    VLAgent +parser+cap+ensemble
      & \makecell{\textbf{61.9\%}\\\gainnaive{+7.5\%}}
      & \makecell{\textbf{58.7\%}\\\gainnaive{+4.6\%}}
      & \makecell{\textbf{60.7\%}\\\gainnaive{+5.5\%}}
      & \makecell{\textbf{60.4\%}\\\gainnaive{+5.7\%}} \\
    \bottomrule
  \end{tabular}%
  }
  \vspace{2pt}
  \label{tab:GQA_ablation}%
  \vspace{-4pt}
\end{table}%
\subsection{Performance Comparison by Visualization}
\label{visualization} 
\begin{figure*}[h!]
    \centering
    \includegraphics[width=0.8\linewidth]{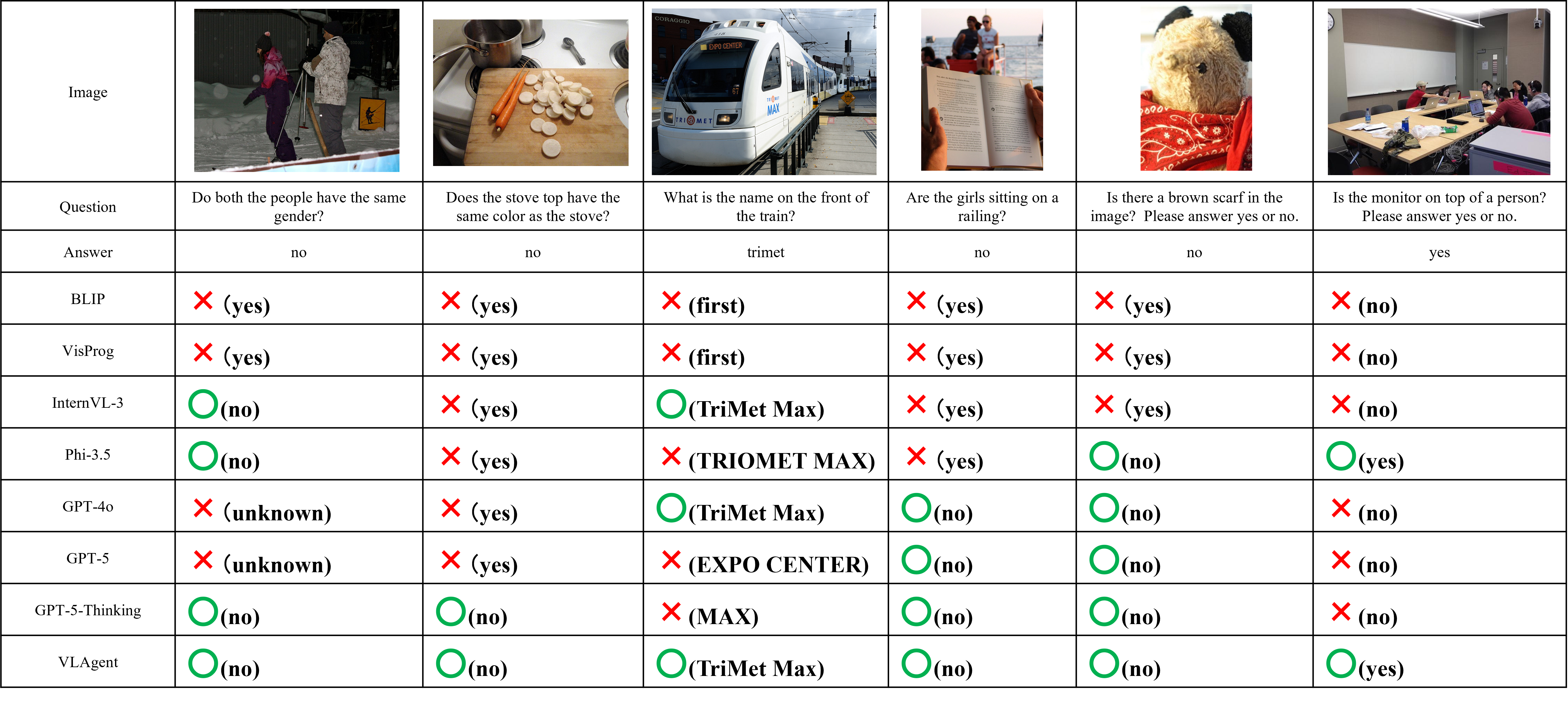}
    \vspace{-12pt}
    \caption{{\small Visual comparison on six image QA examples}}
    \label{fig:five_cases}
    \vspace{-10pt}
\end{figure*}

\textbf{Figure~\ref{fig:five_cases}} illustrates the comparison results of \textbf{Table~\ref{tab:comparison}} with two examples per benchmark from GQA (columns 2~3), VQAv2 (columns 4~5) and MME (columns 6~7). 
%%in single image QA, i.e., two GQA cases, two VQAv2 cases and two MME cases, 
Each example gives a non-trivial text-visual reasoning query. In all six cases, standard VQA models (incl. ensemble VQA method) fail to produce the correct answers, exposing their limitations in performing compositional visual reasoning tasks. Even with a massive amount of training data,  GPT-4o failed on 3 out of the 6 queries. GPT-5 failed on 4 out of the 6 queries. Even GPT-5-Thinking only achieves good performance in 4 out of 6 cases. 
Consider the query in Column 7, the visual image shows clearly where the monitor is. However, GPT-5-Thinking failed on this visual reasoning. In comparison, VLAgent succeeds on all 6 cases by only utilizing lightweight pretrained models, empowered by its neuro-symbolic modularity design for robust compositional visual reasoning.
\begin{figure*} [h!]
    \centering
    \includegraphics[width=0.8\linewidth]{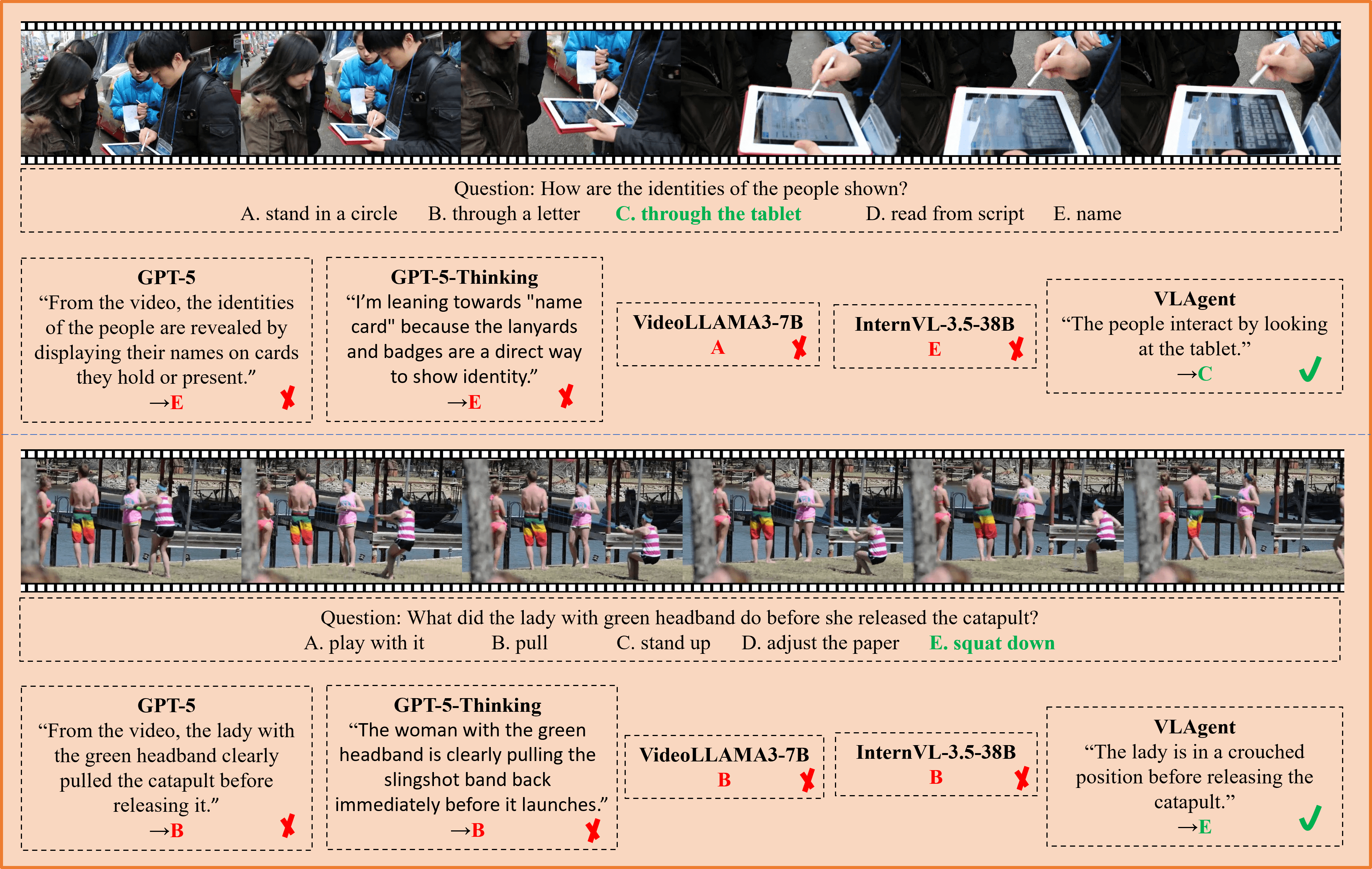}
    \vspace{-8pt}
    \caption{{\small Two video QA examples comparing VLAgent with four selected approach in detail.}}
    \label{fig:two_example}
    \vspace{-13pt}
\end{figure*}
\textbf{Figure~\ref{fig:two_example}} illustrate the effectiveness of VLAgent with two VideoQA examples in NeXT-QA.
%%in the Introduction (\textbf{Figure~\ref{fig:five_cases_nextqa}}) to further illustrate the effectiveness of VLAgent with VideoQA examples. Here, 
For each example, we provide the question, choices, answer, and output to compare VLAgent with 4 STOA methods. For Example-1 (top 2-rows), VLAgent succeeds 
%%in asking more questions about the image and 
%%. Since people are interacting with each other using a tablet and the other four choices are irrelevant in both text query and video (a sequence of video frames), VLAgent is able to 
by conducting compositional visual reasoning to make the correct choice. In comparison, the other four models (GPT-5. GPT-5-Thinking, VideoLLAMA3-7B, InternVL-3.5-38B) all failed to find visual clues to produce the correct inference results. 
%%because they failed to perform compositional visual reasoning to find useful visual clues. 
%%and fails to answer correctly. 
For Example-2 (bottom 2-rows), VLAgent is able to make the correct choice of E (squat down). However, the other four models (GPT-5. GPT-5-Thinking, VideoLLAMA3-7B, InternVL-3.5-38B) all made the wrong choice B (pull) due to the failure to capture/identify the squatting body position. 
%% One possible analysis is that the four models simply capture the coarse movement of the lady wearing the green headband, i.e., pulling the catapult, but all failed to capture/identify the squatting body position right before she releases the catapult. These two examples further illustrate with visualization the effectiveness of putting the four core components of VLAgent (LLM-planner, SS-parser, Script Executor, and Output verifiers) in concert. In \textbf{Figure~\ref{fig:video_example}} of Appendix~\ref{sec:additional_eg}, we provide an example to illustrate how our VLAgent works.
%%Therefore, they all select the wrong choice B (pull). On the contrary, noting the lady is squatting down, VLAgent wins by selecting the correct choice E (squat down).

%Recall \textbf{Figure~\ref{fig:five_cases_nextqa}}, VLAgent failed to perform correct compositional reasoning on the example in column 6 (the last one on the right). Although all other eight VLM models also failed, it is a representative case where VLAgent could further improve its compositional visual reasoning capability. In Appendix~\ref{sec:additional_eg}, we provide more visualization examples, and in appendix~\ref{sec:failure_cases}, we provide  additional failure cases that can serve as a direction of future work.

% \input{sec/related}
\section{Conclusion}
We have presented VLAgent, a neurosymbolic approach to developing a visual-language agent system for compositional visual reasoning with three original contributions: (1) A novel two-stage neurosymbolic architectural design of VLAgent. (2) The use of SS-parser to empower VLAgent to detect and correct logic errors in the LLM-generated planning script. (3) The output evaluation via caption verifier and ensemble verifier to fortify the generalization performance of compositional reasoning. Extensive experiments conducted on 6 visual benchmarks show the effectivess of VLAgent in comparison with over 30 representative VLM$/$ZS methods.

\bibliography{main}
\bibliographystyle{iclr2026_conference}

\clearpage
\appendix
\section*{Appendix Contents}
\begingroup
\hypersetup{linkcolor=black}
\textbf{A. Statements} \hfill \hyperref[sec:stmt]{~\pageref*{sec:stmt}}

\hspace*{1em}A.1 Reproducibility Statement \dotfill \hyperref[sec:repro]{~\pageref*{sec:repro}}

\hspace*{1em}A.2 Large Language Models (LLMs) Usage Statement \dotfill \hyperref[sec:llm_usage]{~\pageref*{sec:llm_usage}}\\

\textbf{B. Additional Examples with Visualization} \hfill \hyperref[sec:additional_eg]{~\pageref*{sec:additional_eg}}

\hspace*{1em}Figure 7: VLAgent Example on Video QA \dotfill \hyperref[fig:video_example]{~\pageref*{fig:video_example}}

\hspace*{1em}Figure 8: Ensemble Verifier Example \dotfill \hyperref[fig:cutting_board_example]{~\pageref*{fig:cutting_board_example}}

\hspace*{1em}Figure 9: Failure Case Due to Insufficient Attribute Check \dotfill \hyperref[fig:bus_example]{~\pageref*{fig:bus_example}}

\hspace*{1em}Figure 10: Failure Case Due to Blur Image \dotfill \hyperref[fig:blur_example]{~\pageref*{fig:blur_example}}\\

\textbf{C. Detailed Experimental Settings} \hfill \hyperref[sec:settings]{~\pageref*{sec:settings}}\\

\textbf{D. Additional Ablation of VLAgent} \hfill \hyperref[sec:add_abl]{~\pageref*{sec:add_abl}}\\

\hspace*{1em}D.1 Ablation of Task Planner LLM \dotfill \hyperref[sec:ablation_llm]{~\pageref*{sec:ablation_llm}}

\hspace*{1em}D.2 Latency Test \dotfill \hyperref[sec:latency]{~\pageref*{sec:latency}}\\

\textbf{E. Implementation Detail} \hfill \hyperref[sec:impl]{~\pageref*{sec:impl}}

\hspace*{1em}E.1 Script Parser \& Script Auditor \dotfill \hyperref[sec:parser_auditor]{~\pageref*{sec:parser_auditor}}

\hspace*{1em}E.2 Ensemble Pruning \dotfill \hyperref[sec:pruning]{~\pageref*{sec:pruning}}\\

\textbf{F. Related Works} \hfill \hyperref[sec:related]{~\pageref*{sec:related}}\\
\bigskip
\endgroup
\newpage
\section{Statements}
\label{sec:stmt}
\subsection{Reproducibility Statement}
\label{sec:repro}
We make every effort to ensure the results in this paper are reproducible.
\begin{itemize}
    \item We provide a link of anonymous GitHub repository where the source code and runtime logs of VLAgent can be downloaded from.
    \item We provide details of modules used in VLAgent in Figure~\ref{fig:modules}. In Appendix~\ref{sec:settings}, we provide the detailed settings of datasets and LLMs. In Appendix~\ref{sec:impl}, we provide full implementation details.
    \item Both in main paper and appendix, we provide figures as examples containing input, procedure and output to show how exactly VLAgent works.
\end{itemize}
\subsection{Large Language Models (LLMs) Usage Statement}
\label{sec:llm_usage}
During the process of writing this paper, LLMs are used and only used for grammar checking and polishing of certain paragraphs.
\section{Additional Examples with Visualization}
\label{sec:additional_eg}
In this supplementary section, we provide additional examples with visualization to illustrate the main optimizations introduced by VLAgent in its backend engine, such as SS-Parser, Per-instruction output verifier via Caption Analysis and Ensemble based Visual Reasoning. 

We first present an example video QA task, where the long video optimization and caption analysis plays the most important role in producing a correct answer. Then we use one example of image-based visual reasoning task to show how ensemble verifier contributes to the overall performance. 
In addition, we also include two failure cases as well as our analysis. 

\begin{figure*} [h!]
    \centering
    \includegraphics[width=0.8\linewidth]{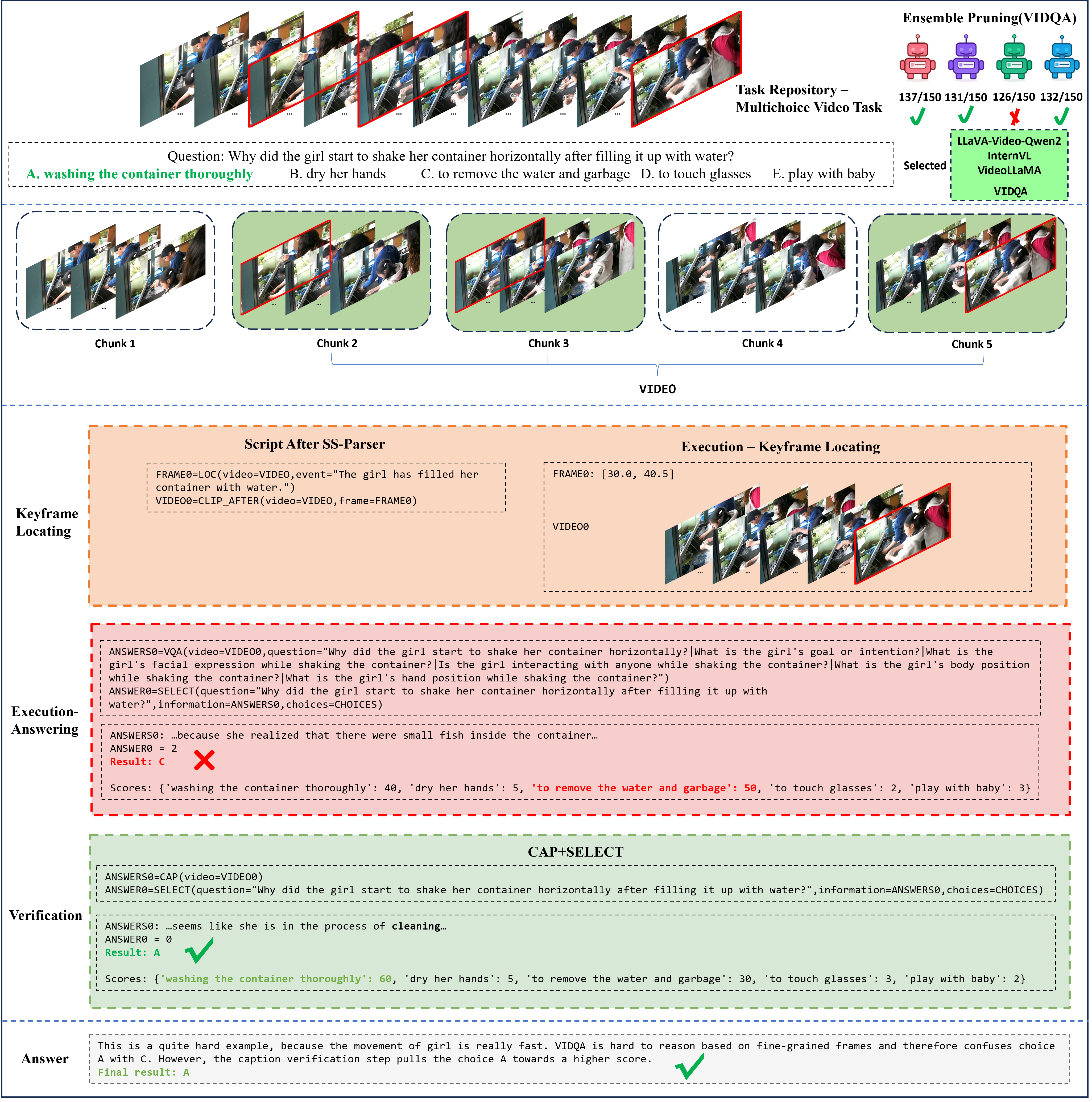}
    \vspace{-4pt}
    \caption{{\small VLAgent example on video QA.}}
    \label{fig:video_example}
    \vspace{-8pt}
\end{figure*}

\textbf{Figure~\ref{fig:video_example}} illustrates a representative text-video reasoning task with the query ``{\em Why did the girl start to shake her container horizontally after filling it up with water?}". The ground truth provided by NeXT-QA is given below the query with A as the correct answer (highlighted in green) out of the five multiple choice answers. 
%%where VLAgent solves an extremely hard problem. 
In this example, the movement of the girl is fast, such that we can only infer it from the overall movement in the entire video. We use the four VLMs smaller than 8B in Table~\ref{tab:nextqa_results} to construct the model pool of the \texttt{VIDQA} module, which is used to answer questions based on a video. During ensemble pruning of the video QA task, we run VLAgent on 150 samples from the task dataset, and run the ensemble pruning process in Algorithm~\ref{alg:model_selection} to select three models. Finally, LLaVA-Video-7B-Qwen2~\cite{zhang2024videoinstructiontuningsynthetic}, InternVL-3.5-8B~\cite{wang2025internvl3_5} and VideoLLaMA3-7B~\cite{damonlpsg2025videollama3} are selected to construct the \texttt{VIDQA} module. Since the original video is too long, we divide the video into 5 chunks, with each chunk containing roughly 15 seconds, and use LLaVA-Video-7B-Qwen2~\cite{zhang2024videoinstructiontuningsynthetic} to get a detailed caption of each chunk. Then GPT-4.1-mini~\cite{openai2025gpt41} is used to select chunks which may be relavent to the question. Chunks 2, 3 and 5 are selected, and they are concatenated together to form variable \texttt{VIDEO}. Meanwhile, our task planner generates a script to solve the problem. The script is then run with an initial variable \texttt{VIDEO}. In the figure, we divide the script execution process into two stages: keyframe locating and question answering. In the first stage, \texttt{LOC} is called to locate the timestamp where the girl has filled her container with water, and then clips the video after the starting frame. The clipped video is shown on the right of \texttt{VIDEO0} in Figure~\ref{fig:video_example}, with keyframes annotated with red border. We can already see the girl is washing the container. In question answering stage, \texttt{VIDQA} is called to ask a bunch of questions on \texttt{VIDEO0} and \texttt{SELECT} is called to choose the best answer. Unfortunately, video QA models do not provide enough information to distinguish A with C, and mistakenly votes C as the most possible answer. In verification step, we use per-frame caption model on \texttt{VIDEO1}, and answer A is clearly supported. Therefore, the answer is modified to answer A and VLAgent generates the correct final answer to this question.

\begin{figure*}[h!]
    \centering
    \includegraphics[width=0.9\linewidth]{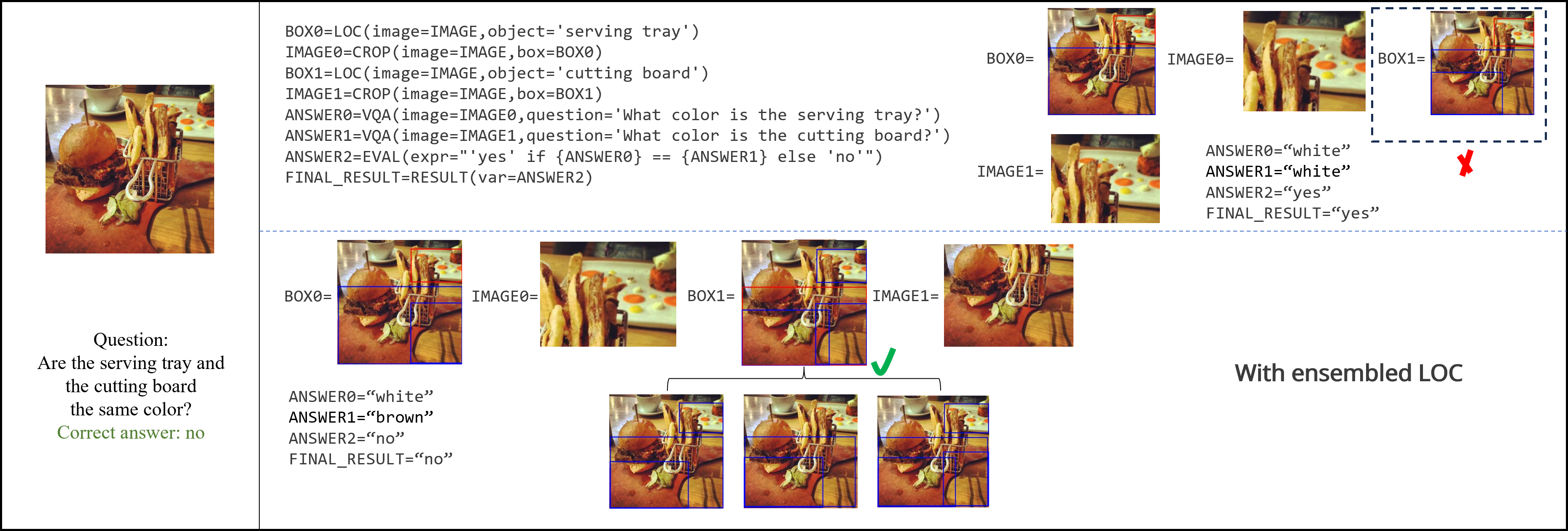}
    \caption{A GQA example illustrates the effectiveness of ensemble verification in VLAgent.}
    %%, because all three models vote for that bounding box with a high confidence score.}
    \label{fig:cutting_board_example}
\end{figure*}

% \textbf{Figure~\ref{fig:baseline1_example_deer}} shows an example from the GQA benchmark~\cite{hudson2019gqa} where our script parser and script auditor detect and correct the logic errors produced by the LLM-generated planning script. In the third line, our SS-Parser detects an invalid ``object" parameter, i.e., ``standing", which is impossible for object detection models to detect. To fix the issue, we can either modify the parameter as ``standing deer", or directly use VQA. In our current implementation, we directly replace the script as a VQA call because detecting "standing deer" requires a highly accurate object detection model, otherwise it is prone to make false positive detections (e.g., detecting a sitting animal as standing).

\textbf{Figure~\ref{fig:cutting_board_example}} illustrates the effect of ensemble-verifier by an example from GQA with visualization. It shows that with ensemble verification, VLAgent can further improve the LOC performance using three external object detection models chosen by our ensemble-verifier. The cutting board bounding box in BOX1 gets the highest ensemble confidence score based on inconsistency resolution and fusion analysis. As we can see from the figure, the script locates the serving tray and cutting board, and then queries and compares their colors, which is correct. However, the default \texttt{LOC} module in VLAgent returns a wrong bounding box when it performs inference to locate the cutting board, making \texttt{IMAGE1} remains to be a serving tray, and thus outputs a wrong result. By leveraging two other \texttt{LOC} modules, VLAgent can verify the output of the original \texttt{LOC} module. As shown in the bottom of Figure~\ref{fig:cutting_board_example}, both of the additional \texttt{LOC} modules can  successfully locate the cutting board. 
%%and only one of them mistakenly recognizes the serving tray as the cutting board. 
As a result, the cutting board is the top-1 bounding box ranked by the fusion confidence score instead of the serving tray, delivering the correct final result by VLAgent with ensemble boosting.

\begin{figure*} [h!]
    \centering
    \includegraphics[width=\linewidth]{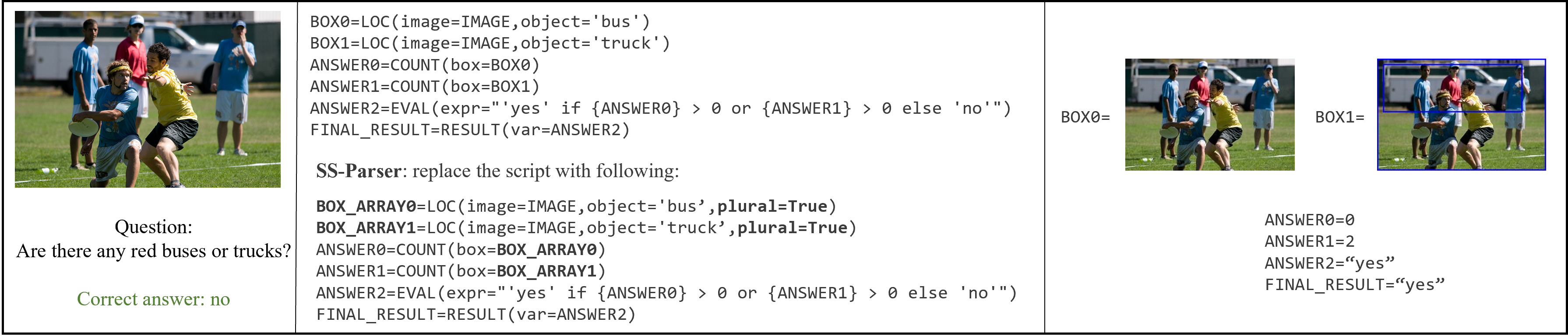}
    \caption{The script should examine the color of buses/trucks, but the SS-Parser fails to detect this type of error.}
    \label{fig:bus_example}
\end{figure*}

\begin{figure*} [h!]
    \centering
    \includegraphics[width=\linewidth]{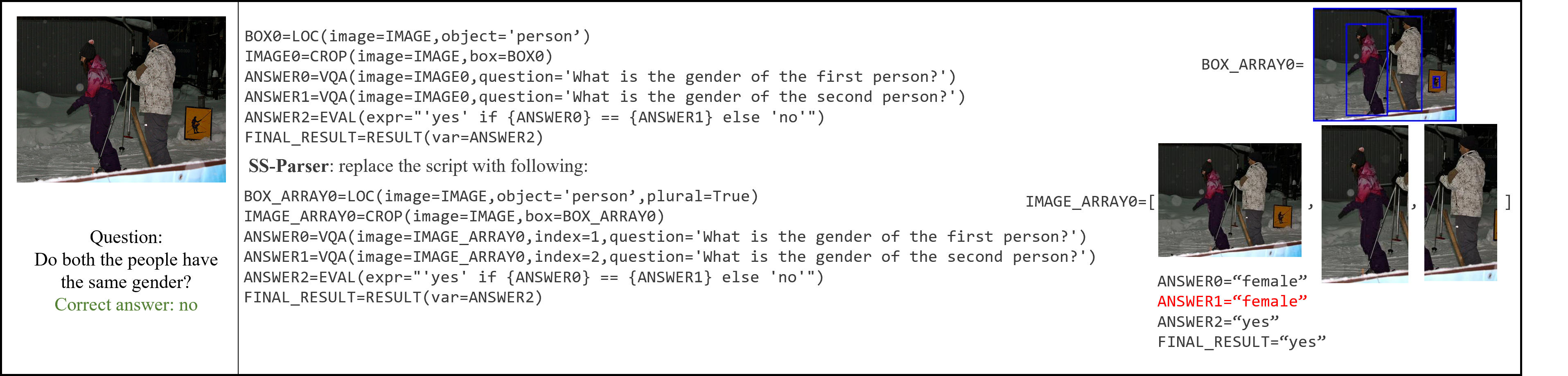}
    \caption{An example where VLAgent fails because the image is blurred.}
    \label{fig:blur_example}
\end{figure*}

We also provide two failure scenarios from GQA in \textbf{Figure~\ref{fig:bus_example}}, and \textbf{Figure~\ref{fig:blur_example}}. 

In \textbf{Figure~\ref{fig:bus_example}}, the script neglects to examine the color of the buses and trucks because the 20 in-context examples do not cover similar questions. Consequently, the SS-Parser fails to detect this type of logical error. This failure case motivates our future work in developing a more advanced semantic parser. For example, our revised SS-Parser can detect the error by judging the adjective word "red" in the question is not checked in the script. 

Finally, in \textbf{Figure~\ref{fig:blur_example}}, the VLAgent successfully corrects the errors in LLM-generated planning script via its SS-Parser and script-repair module. Unfortunately although the ensemble reasoning is leveraged, it fails when all of the three VQA models produce incorrect answers for the second person due to the fact that the image is too blurred and dark. This indicates that a more robust output verification method would improve the solution.
\section{Detailed Experimental Settings}
\label{sec:settings}
In our experiments, we evaluate the performance of VLAgent and the baselines on six distinct datasets, each corresponding to a specific task. The datasets are described below:

\begin{itemize}
    \item \textbf{GQA}~\cite{hudson2019gqa}: A large-scale dataset for real-world visual reasoning and compositional question answering. GQA challenges models to understand complex scenes and answer questions that require multi-step reasoning. For example, a question might ask whether there is a boy to the left of a standing girl. For fair comparison, following the Visprog settings, we test VLAgent on a subset of the test\_dev set by randomly selecting 20 questions per type, resulting in 1460 QA pairs.
    \item \textbf{NLVR2}~\cite{suhr2018corpus}: A dataset designed for natural language visual reasoning, consisting of paired images and textual statements. The task is to determine whether a statement accurately describes the visual content, thereby assessing both language understanding and visual grounding. We adopt the same setting as Visprog by evaluating on the entire balanced test set. After filtering out expired image links, 2316 samples remain.
    \item \textbf{VQAv2}~\cite{goyal2017making}: An improved version of the original VQA~\cite{antol2015vqa} dataset, VQAv2 provides balanced question-answer pairs to mitigate language biases and foster deeper visual understanding. Widely used for benchmarking visual question answering models, we randomly sample 20 QA pairs per type from the validation set (the test set is not used as it lacks answer annotations). With 65 question types, this yields a total of 1300 QA pairs.
    \item \textbf{MME}~\cite{fu2024mmecomprehensiveevaluationbenchmark}: A comprehensive benchmark designed to evaluate various aspects of multimodal reasoning, including math reasoning, code understanding, chart QA, existence, position, color, and more. In this work, we focus on a representative subset of the dataset—specifically the existence, position, and color categories—which require direct image-level reasoning. This selection aligns with our objective of evaluating visual reasoning capabilities independent of external components such as OCR or symbolic math engines. Categories heavily reliant on textual extraction or LLM-based arithmetic fall outside the scope of our image-centric evaluation. Our subset includes 580 QA pairs sampled from the official test set, providing a robust and targeted assessment of image-level reasoning performance.
    \item \textbf{NeXT-QA}~\cite{xiao2021next}: A large-scale \emph{video} question answering benchmark targeting \emph{temporal} and \emph{causal} reasoning about human activities and events. Questions are multiple-choice and often require cross-frame understanding of before/after relations, intentions, and cause/effect. We evaluate on a subset of its test set, where we randomly sample 200 QAs per type, forming a subset of 1493 QAs.

    \item \textbf{HC-RefLOCO}~\cite{10.5555/3737916.3740138}: A human-centric referring expression comprehension dataset with \emph{long}, attribute-rich descriptions that combine appearance, pose, spatial relations, and actions. The task is to localize the target person (bounding box) that best matches the natural-language description, emphasizing fine-grained grounding over long sentences. We sample 1400 referring expressions from the official test split to form a subset to test VLAgent and other approaches.

\end{itemize}
Our experiments are conducted on a single H100 or H200 GPU in a Python 3.9 environment. We evaluate four LLM models for script generation:

\begin{itemize}
    \item \textbf{GPT-3.5}~\cite{openai2023gpt35turboinstruct}: \texttt{gpt-3.5-turbo-instruct} API from OpenAI is employed to assess GPT's performance. For fair comparison, we change the LLM models of zeroshot baselines to GPT-3.5 as well if they are using older models.
    \item \textbf{Llama3-8B}~\cite{grattafiori2024llama3herdmodels}: This model is loaded from HuggingFace at \texttt{meta-llama/Meta-Llama-3-8B}. Developed by Meta, it is a completion model with 8B parameters, a vocabulary size of 128K, and a context window of 8K - meeting the basic requirements for script generation.
    \item \textbf{GLM4-9B}~\cite{glm2024chatglm}: This model is loaded from HuggingFace at \texttt{THUDM/glm-4-9b-hf}. It is a more advanced model, featuring 9B parameters, a vocabulary size of 152K, and a context window of 128K.
    \item \textbf{Mistral-Small-24B-Base-2501}~\cite{jiang2023mistral7b}: From HuggingFace at \texttt{mistralai/Mistral-Small-24B-Base-2501}, this model is loaded. Among the four LLMs, Mistral offers the best performance, with a parameter size of 24B (even larger than GPT-3.5, which has a reported parameter size of 20B~\cite{singh2023codefusion}). Its vocabulary size is 131K, and its context window is 32K.
\end{itemize}

% \vspace{48pt}
\section{Additional Ablation of VLAgent}
\label{sec:add_abl}
\subsection{Ablation of Task Planner LLM}
\label{sec:ablation_llm}

{\bf Table~\ref{tab:main_result}} reports the comparison results of full-fledged VLAgent with its naive version with only LLM-generated script and its runtime executor (w/o parsers and verifiers) with four popular LLMs as the LLM-script generator respectively and tested on all four ImageQA benchmarks. VLAgent consistently outperforms its naive version by a significant margin. In particular, for NLVR2 with GLM, the combo performs poorly with only an accuracy of 19.2\% due to incorrect generation of LLM programs. In comparison, VLAgent achieves an accuracy of 58.6\% with 39.4\% gain margin. Similarly, for VQAv2, VLAgent significantly improves the accuracy with Llama by 14.3\%, Mistral by 6.4\%, GPT-3.5 by 4.6\%, and GLM by 3.9\%. For MME, the full-fledged VLAgent remains the top performer with 3.5\% improvement on average. Given that the questions in the selected categories of MME are significantly simpler compared to those in GQA, the naive VLAgent can achieve very good accuracy. The improvement brought by the combo of SS-Parser and Output verifier is about $1.4 \sim 7.6$\%. 
\vspace{-10pt}
\begin{table*}[h!]
  \centering
  \caption{{\small Accuracy Comparison on 4 benchmarks (GQA, VQAv2, MME, NLVR2). Each benchmark is tested using four different LLMs as the corresponding initial task plan generators for both VLAgent naive (with only LLM script planner and executor) and VLAgent (the full fledged version with SS-Parser and caption and ensemble Verifiers). A total of 16 combos for VLAgent to compare with 16 combos of VLAgent naive, showing the consistent gain of SS-Parser and Output Verifiers. 
  }}
  {\small
    \begin{tabular}{c|c|c|c|c|c}
    \toprule
    Agent Framework & Benchmark  & GPT 3.5   & Llama & Mistral & GLM \\
    \midrule
    VLAgent naive & GQA & 54.4\% & 54.1\% & 55.2\% & 54.7\% \\
    VLAgent & GQA & {\bf 61.9\%} & {\bf 58.7\%} & {\bf 60.7\%} & {\bf 60.4\%} \\ 
    Improvement & GQA & {\bf \textcolor{green!70!black}{+7.5\%}} & {\bf \textcolor{green!70!black}{+4.6\%}} & {\bf \textcolor{green!70!black}{+5.5\%}} & {\bf \textcolor{green!70!black}{+5.7\%}} \\ 
      \midrule
    VLAgent naive & VQAv2 & 72.3\% & 61.0\% & 70.8\% & 74.2\% \\
    VLAgent & VQAv2 & {\bf 76.9\%} & {\bf 75.3\%} & {\bf 77.2\%} & {\bf 78.1\%} \\
    Improvement & VQAv2 & {\bf \textcolor{green!70!black}{+4.6\%}} & {\bf \textcolor{green!70!black}{+14.3\%}} & {\bf \textcolor{green!70!black}{+6.4\%}} & {\bf \textcolor{green!70!black}{+3.9\%}} \\ 
      \midrule
    VLAgent naive & MME & 86.9\% & 81.0\% & 85.3\% & 86.0\% \\
    VLAgent & MME & {\bf 88.4\%} & {\bf 88.6\%} & {\bf 88.8\%} & {\bf 87.4\%} \\
    Improvement & MME & {\bf \textcolor{green!70!black}{+1.5\%}} & {\bf \textcolor{green!70!black}{+7.6\%}} & {\bf \textcolor{green!70!black}{+3.5\%}} & {\bf \textcolor{green!70!black}{+1.4\%}} \\
      \midrule
    VLAgent naive & NLVR2 & 69.3\% & 65.9\% & 67.6\% & 19.2\% \\
    VLAgent  & NLVR2 & {\bf 73.2\%} & {\bf 70.0\%} & {\bf 74.1\%} & {\bf 58.6\% } \\
    Improvement & NLVR2 & {\bf \textcolor{green!70!black}{+3.9\%}} & {\bf \textcolor{green!70!black}{+4.1\%}} & {\bf \textcolor{green!70!black}{+6.5\%}} & {\bf \textcolor{green!70!black}{+39.4\%}} \\
    \bottomrule
    \end{tabular}%
    }
     \vspace{-12pt}
  \label{tab:main_result}%
\end{table*}%
\subsection{Latency Test}
\label{sec:latency}
The third experiment we want to report as a part of the ablation study is the latency of each core component and each optimization, especially the cost of ensemble verifier and caption verifier. 
%%We also conduct an experiment on GQA dataset to test how each component adds the lagency. 
{\bf Table~\ref{tab:latency}} reports the per-sample inference time of VLAgent on GQA, including its naive version (planner+executor only), and the individual core components within VLAgent. It is observed that empowered with all the SS-parser checking, repairing, and verifying mechanisms, VLAgent runs at 7.54 seconds per sample. Compared to the VLAgent naive which takes 3.24 seconds, the full-fledged VLAgent offers the inference latency at an acceptable range in practice.
Also the breakdown of the components reveals that a majority of the added latency stems from the caption verifier, which invokes both an image captioning model and an LLM. In comparison, the ensemble fusion of vision models introduces only a modest overhead and the SS-Parser incurs negligible additional cost.
\vspace{-15pt}
\begin{table}[h!]
  \centering
  \caption{Inference time per sample. VLAgent w. parallel means part of the caption verifier runs in parallel with the other VLAgent components.}
  %%task planner, SS-Parser, plan repairer, output verifier, plan executor.}
    \begin{tabular}{c|c}
    \toprule
    Component & Time Cost (s) \\
    \midrule
   %% Visprog 
    VLAgent Naive & 3.24 \\
    VLAgent (all) Sequential & 7.54 \\
    {\bf VLAgent (all) Parallel} & \textbf{5.10} \\
     \midrule
    Task Planning & 1.50 \\
    LOC w/o ensemble verifier & 0.23 \\
    LOC w/ ensemble verifier & 0.88 \\
    VQA w/o ensemble verifier & 0.17 \\
    VQA w/ ensemble verifier & 0.23 \\
    SS-Parser & 0.00 (0.0016) \\
    Caption Verifier & 4.01 \\
    \bottomrule
    \end{tabular}%
  \label{tab:latency}%
\end{table}%

Important to note is that the captioning and its analysis are independent of script generation and script execution, we can process them in parallel. For parallel implementation of caption-verifier, the total inference time of VLAgent can be reduced to 5.10 seconds per sample (a gain margin of 2.44 seconds). This also indicates that the checking and verifying mechanism only adds a tiny latency increase (less than 2s), while offering substantially improved visual compositional reasoning capability and improved interpretability.
\section{Implementation Detail}
\label{sec:impl}
\subsection{Script Parser \& Script Auditor}
\label{sec:parser_auditor}
Before passing the script to the executor, VLAgent checks for potential errors. The \emph{script parser} looks for syntax errors; for example, it flags any script that uses a module name not supported by VLAgent. The \emph{script auditor} checks for semantic-level errors. For instance, if the script attempts to locate “standing” in a cropped bird image to check whether a bird is standing, the auditor will note that the object name passed to the \texttt{LOC} module should be a noun or a phrase describing a noun. The auditor can detect such an error and return a corrected script.

\textbf{Table~\ref{tab:error_types}} summarizes the representative \emph{detected conditions} and the \emph{automatic repairs} applied by our two safeguards: the \emph{script parser} and \emph{script auditor}. For the script auditor, an expression like \verb|== 'yes'| must be replaced with \verb|== True| because we omit explicit type conversion in the template for simplicity. Instead, all variables are converted to their appropriate types inside the \texttt{EVAL} module: digit strings become numbers, “yes” and “no” become \texttt{True} or \texttt{False}, and other formats remain unchanged. However, the LLM is unaware of this and may still produce statements like \verb|{ANSWER0} == 'yes'| to determine if a variable is “yes” or “no.”

The script auditor may also add a \texttt{plural=True} flag in the \texttt{LOC} call if the object name is plural or corresponds to a plural word in the question. For example, if the question is “Are both people the same gender?” and \texttt{LOC} identifies a person, the auditor can detect that “person” corresponds to “people,” meaning the bounding box should encompass a group of people. In such a case, the script auditor adds \texttt{plural=True} to the \texttt{LOC} call, and the bounding box of the entire image is returned if any person is detected. In implementation, the script parser and the script auditor can be implemented together, with a line-by-line check of the script, as shown in Algorithm~\ref{alg:refined_parser}.
\vspace{-10pt}
\begin{table*}[htbp]
  \centering
  \caption{Representative issues and automatic repairs applied by the script parser and script auditor.}
  {\small
    \begin{tabular}{c|c|c}
    \toprule
    \textbf{Module} & \textbf{Detected Condition} & \textbf{Strategy} \\
    \midrule
    \multirow{4}[2]{*}{Script parser} 
      & Wrong script format & Replace with direct VQA call \\
      & Non-existent module names & Replace with direct VQA call \\
      & Non-existent variables & Replace with direct VQA call \\
      & Syntax error in EVAL's expression & Replace with direct VQA call \\
    \midrule
    \multirow{4}[2]{*}{Script Auditor} 
      & LOC: Strange object names & Replace with direct VQA call \\
      & LOC: plural object name & Add ``plural=True`` in LOC call \\
      & LOC: corresponding plural noun in question found & Add ``plural=True`` in LOC call \\
      & EVAL contains ``== 'yes'`` & Replace it with ``== True`` \\
      & EVAL contains ``== 'no'`` & Replace it with ``== False`` \\
    \bottomrule
    \end{tabular}%
    }
  \label{tab:error_types}%
\end{table*}%

\begin{algorithm}[t]
\caption{High-Level Algorithm for SS-Parser. For simplicity, this algorithm only includes modules in GQA.}
\label{alg:refined_parser}
\DontPrintSemicolon
\KwIn{program: multiline script;\quad
      question: user query;\quad
      module\_list: allowed module names;\quad
      var\_dict: runtime variable values (e.g.\ \{\texttt{"IMAGE"}:None\})}
\KwOut{approved\_program: adjusted script or fallback}
\Begin{
  Split \texttt{program} into \texttt{lines};  
  Set \texttt{num\_box\_arrays}=0 and \texttt{num\_image\_arrays}=0;  
  \ForEach{(i, $\ell$) in \texttt{enumerate(lines)}}{
    Parse $\ell$ with \texttt{parse\_step} into $(s,o,a)$;  
    \If{$s$ not in \texttt{module\_list}}{\Return fallback\_script;}  
    \Switch{$s$}{
      \Case{EVAL}{
        Let \textit{expr\_fmt} = \texttt{a["expr"]};\\  
        \If{\textit{expr\_fmt} syntax error found in \textit{expr\_fmt}}{\Return fallback\_script;}  
        Set \texttt{var\_dict[o]} = \texttt{eval(expr\_fmt)};  
        In $\ell$, replace \texttt{== 'yes'}→\texttt{== True} and \texttt{== 'no'}→\texttt{== False};  
      }
      \Case{LOC}{
      Set \texttt{var\_dict[o]} = [[0,0,100,100]];\\
      \If{\texttt{a["image"]} not in \texttt{var\_dict}}{\Return fallback\_script;}
        Let \textit{obj} = \texttt{eval(a["object"])};\\ 
        \If{obj is not noun OR obj not mentioned in question}{\Return fallback\_script;}  
        \If{obj is plural or corresponds to a plural noun in question}{\texttt{a["plural"]=True};\\
        \If{\texttt{question} contains any of \{\texttt{all}, \texttt{every}, \texttt{both}, \texttt{each}\}}{
        $k\leftarrow$\texttt{num\_box\_arrays}++;\\
        $o\leftarrow$``BOX\_ARRAY\_\{k\}";\\
          For each CROP line using the bounding box: update bounding box name as new $o$; for each related VQA call: $m\leftarrow$\texttt{num\_image\_arrays}++, update image argument name to be \texttt{IMAGE\_ARRAY\_m}, add increasing index argument starting from 1;
        }}  
      }
      \Case{VQA}{
        \If{\texttt{a["image"]} not in \texttt{var\_dict}}{\Return fallback\_script;}  
        Set \texttt{var\_dict[o]} = '0';  
      }
      \Case{CROP/COUNT/RESULT/GET}{
        Run $l$, updating \texttt{var\_dict};  
        \If{exception occurs}{\Return fallback\_script;}   
      }
      \Else{%
      \Return fallback\_script;
    }%
    }% end switch s
    Replace line i of \texttt{lines} with the updated $\ell$;  
  }% end for
  Let approved\_program = join(lines, "\textbackslash n");\\
  \Return approved\_program;
}
\end{algorithm}
\setlength{\textfloatsep}{0pt}

On video QA task, SS-Parser adds the syntax check for \texttt{VIDQA}, \texttt{LENGTH}, \texttt{CLIP}, \texttt{CLIP\_AROUND}, \texttt{CLIP\_BEFORE}, \texttt{CLIP\_AFTER}, \texttt{SELECT} modules, while for \texttt{LOC}, the input is changed to question along with a video. The script auditing part extracts the temporal keywords from the question and checks whether the usage of \texttt{CLIP} family models is correct. For referring expression, SS-Parser only does syntax check on supported modules.
%\subsection{Core Modules}
%\label{sec:external_modules}
%Figure~\ref{fig:modules} shows a complete list of core modules used in VLAgent. The module names are the naming convention learned by the Planner via a combo of $K$-shots and CoT for in-context instruction-learning. The first six rows are external pre-trained models marked in blue. The remaining modules marked in green, each corresponds to an internal module in the VLAgent Python library. \texttt{LOC}, \texttt{VQA}, \texttt{CAP}, \texttt{CROP} to \texttt{GET}, and \texttt{RESULT} are used for GQA, VQAv2, MME workloads.
%\begin{figure*} [h!]
%    \centering
%    \includegraphics[width=0.85\linewidth]{figs/modules.png}
%    \vspace{-8pt}
%    \caption{{\small Core Modules supported in VLAgent. 
    %\textcolor{blue}{
%    For NLVR2 task, only \texttt{VQA}, \texttt{EVAL} and \texttt{RESULT} are used. Video QA task uses modules accepting video as input plus \texttt{SELECT} and \texttt{EVAL}. HC-Ref-LOCO videoQA task uses \texttt{LOC}, \texttt{CAP}, \texttt{FIND}, and \texttt{VOTE}.}}
%    \label{fig:modules}
%    \vspace{-10pt}
%\end{figure*}
\subsection{Ensemble Pruning}
\label{sec:pruning}
Consider \texttt{LOC} as an example.
%%to illustrate how we select best set of models. 
Suppose we have $N$ models to consider as candidate external modules. Instead of designing an ensemble of $N$ models, we consider only a small subset of $M$ models by ensemble pruning method~\cite{tekin-etal-2024-llm} to avoid computation overhead. 
%%
%% Selim Tekin, Fatih Ilhan, Sihao Hu, Tiansheng Huang, Margaret Loper, Ling Liu. \href{https://dl.acm.org/doi/10.1145/3746457}{Robust Few-Shot Ensemble Learning with Focal Diversity-Based Pruning}. ACM Transactions in on Intelligent Systems and Technology. 
%%
%%
%%We take $m$ samples from either the training set or test set of the task. There's no risk of data leakage if we use the test set, because we don't leverage the ground truth answers and the data distributions of the training and test set are the same. During the process, we record the results of each \texttt{LOC} step. Suppose the we run $n$ \texttt{LOC} lines in total. For the $i$th data ($1\le i\le n$), we get the bounding box list of $N$ models as $B_{ij}$, where $j=1,2,\cdots,N$. We run any ensemble algorithm and get the final bounding box list $B$, which serves as a \textit{pseudo label}.
%%We run any ensemble algorithm and get the final bounding box list $B$, which serves as a \textit{pseudo label}.

We run our VLAgent on a small sample from the task dataset. Suppose we totally run \texttt{LOC} for $P$ times. Our ensemble pruning step consists of \textbf{three stages}: (i) Compute a score for each model to represent its overall performance; (ii) Use K-Means~\cite{macqueen1967multivariate} to group models with similar performance together. Before clustering, we use Silhouette distance~\cite{rousseeuw1987silhouettes} to compute the best $K$. (iii) The models of the cluster with the highest scores are put to our candidate model list. If the number of candidate models is smaller than our desired $M$, go through (ii) and (iii) on remaining models. 

For each of the $P$ \texttt{LOC} instructions,  we get the bounding box list of $M$ models, denoted as $B_{ij}$, where $1\le i\le P$,  $1\le j\le M$. Let $B$ denote the final list of the $M$ bounding boxes produced by the ensemble fusion of $M$ external  \texttt{LOC} models, , which serves as a \textit{pseudo label}. Let $Area(B):=\{(x,y)|\exists b\in B, (x,y) \in b\}$ denote the union region of bounding boxes in bounding box list $B$.  Hence, we can compute the confidence score $I_{ij}$ for each bounding box list $B_{ij}$ ($1\le i\le P$, $1\le j\le M$), as follows:
\begin{align}
    I_{ij}=\frac{|Area(\cup B_{ij})\cap Area(\cup B)|}{|Area(\cup B_{ij})\cup Area(\cup B)|}
\end{align}
which is the Intersection of Union (IoU) between the $M$ bounding boxes in $B_{ij}$ and pseudo label. $|\cdot|$ is the area. We then use the average IoU of the $j$ th model as its score:
\begin{align}
s_j=\frac{1}{P}\sum_{i=1}^PI_{ij}
\end{align}
After getting $\{s_j\}_{j=1}^N$, we go through stage (ii) and (iii) to iteratively select $M$ models to construct the model set for \texttt{LOC}. Refer to \textbf{Algorithm~\ref{alg:model_selection}} for a detailed selection pseudo code. In Algorithm~\ref{alg:model_selection}, line 2 to line 6 runs the VLAgent on $m$ samples from the task dataset. Line 7-10 goes through stage (i) to get a score of each model. Line 11-22 uses Silhouette distance~\cite{rousseeuw1987silhouettes} to select the best $K$, where $C(s_j)$ is the cluster $s_j$ belongs to, $a(j)$ is the intra-cluster distance for $s_j$, $b(j)$ is the inter-cluster distance for $s_j$, and $s(j)$ is the Silhouette distance of $s_j$. Line 23-26 runs $K*$-Means clustering, and adds the models in cluster $C^*$ which contains largest $s_j$s to the candidate model set $\mathcal{S}$, meanwhile removing them from the model pool $\mathcal{R}$.

\begin{algorithm}[h!]
\caption{Model Selection via IoU-based Scoring and K-Means Clustering}
\label{alg:model_selection}
\DontPrintSemicolon
\KwIn{$N$ models; $m$ samples from dataset; minimum number of models $M$}
\KwOut{$\mathcal{S}$: set of selected model indices}

Initialize $\mathcal{S} \gets \emptyset$, $\mathcal{R} \gets \{1, 2, \ldots, N\}$\;

Sample $m$ data points and run \texttt{LOC} for $P$ total lines\;

\For{$i = 1$ \KwTo $P$}{
    \For{$j \in \mathcal{R}$}{
        Get bounding box list $B_{ij}$ from model $j$\;
    }
    Run ensemble algorithm to get pseudo label $B_i$\;
}

\For{$j \in \mathcal{R}$}{
    \For{$i = 1$ \KwTo $n$}{
        $I_{ij} = \frac{|Area(\cup B_{ij}) \cap Area(\cup B_i)|}{|Area(\cup B_{ij}) \cup Area(\cup B_i)|}$\;
    }
    $s_j = \frac{1}{P}\sum_{i=1}^P I_{ij}$\;
}
\While{$|\mathcal{S}| < M$ and $\mathcal{R} \neq \emptyset$}{
    \For{$K = 1$ \KwTo $|\mathcal{R}|$}{
        Run K-Means clustering on $\{s_j\}_{j \in \mathcal{R}}$ to get clusters $\{C_r\}_{r=1}^K$\;
        \For{$j \in \mathcal{R}$}{
            \eIf{$|C(s_j)| = 1$}{
                $s(j) = 0$\;
            }{
                $a(j) = \frac{1}{|C(s_j)|-1}\sum_{s_k \in C(s_j), k \neq j}|s_j - s_k|$\;
                $b(j) = \min_{r \neq C(s_j)} \frac{1}{|C_r|}\sum_{s_k \in C_r}|s_j - s_k|$\;
                $s(j) = \frac{b(j) - a(j)}{\max\{a(j), b(j)\}}$\;
            }
        }
        $S(K) = \frac{1}{|\mathcal{R}|}\sum_{j \in \mathcal{R}} s(j)$\;
    }
    
    $K^* = \arg\max_{K \in \{1, 2, \ldots, |\mathcal{R}|\}} S(K)$\;
    Run $K^*$-Means clustering on $\{s_j\}_{j \in \mathcal{R}}$\;
    $C^* = \arg\max_{r \in \{1, \ldots, K^*\}} \max_{j: C(s_j) = r} s_j$\;
    $\mathcal{S} \gets \mathcal{S} \cup \{j : C(s_j) = C^*\}$\;
    $\mathcal{R} \gets \mathcal{R} \setminus \{j : C(s_j) = C^*\}$\;
}

\Return{$\mathcal{S}$}\;
\end{algorithm}
\setlength{\textfloatsep}{0pt}

For other external modules like \texttt{VQA}, we run the same algorithm. The only difference is how $I_{ij}$ is computed. For \texttt{VQA}, $I_{ij}=\mathbb{I}[|A_{ij}\cap A_i|>0]$, where $A_{ij}$ is the set of words generated by model $j$ on $i$-th data, and $A_i$ is the set of words in the ensembled answer. For \texttt{VIDQA}, $I_{ij}=\mathbb{I}[choice_{ij} = choice_i]$, where $choice_{ij}$ is the choice supported by model $j$ on $i$-th data, and $choice_i$ is the majority voted choice. The effectiveness of ensemble verifier is measured in the Ablation study in Section~\ref{sec:ablation}.

% Required packages:
% \usepackage{algorithm}
% \usepackage{algorithmic}

% This algorithm uses algorithm2e package only
% Remove these conflicting packages: algpseudocode, algorithm, algorithmic
% Keep only: \usepackage[linesnumbered,ruled,vlined]{algorithm2e}

\section{Related Works}
\label{sec:related}
%\noindent {\bf Modular and Compositional Visual Reasoning.}
Our work is inspired by several pioneering projects in 
%%modular vision for compositional visual reasoning, such as VisProg~\cite{gupta2023visual}, ViperGPT~\cite{suris2023vipergpt}, and 
Neural Module Networks (NMNs)~\cite{andreas2016neural,hu2018explainable,Hu_2017_ICCV,johnson2017inferring}. NMNs were introduced to improve interpretability by decomposing visual reasoning into explicit sub-tasks. In NMN frameworks, a question is parsed into a layout of modular operations (e.g., find, filter, count), each of which is handled by specialized neural units, and the results are composed to produce the answer. This compositional design yields a step-by-step reasoning trace that is more transparent than monolithic end-to-end models. 
%%Classic NMNs demonstrated the benefits of modular reasoning on benchmark datasets like SHAPES~\cite{andreas2016neural} and VQA~\cite{antol2015vqa}, showing that complex queries can be answered by linking simpler learned functions. 
However, NMNs require supervised learning to train the module selection or layout predictor, often relying on ground-truth programs or strong annotations. Consequently, their generalization is constrained by the quality and quantity of training data, and hard to extend to new tasks without additional supervision.
The recent progress in Neural Module Networks includes ViperGPT~\cite{suris2023vipergpt}, VisProg~\cite{gupta2023visual} and GENOME~\cite{chen2023genome}. These recent projects formulate Python programs that invoke external trained models and Python library modules to obtain answers in visual reasoning tasks. 
%%They both generate a program and then use an executor/interpreter to execute it. 
%%The difference is that ViperGPT directly uses Python language and supports more flexible programs, but its program often encounters execution error and is less interpretable. 
%%GENOME~\cite{chen2023genome} is built upon Visprog. It uses a small data split as its "training set." During the training stage, it runs Visprog to perform inference on this training set, and then leverages an LLM to determine whether new external modules are needed. 
%%Finally, GENOME generates the code for the new modules and the corresponding in-context learning examples. 

Our research is also inspired by recent research in LLM enhancement for visual reasoning~\cite{10161317, 
suris2023vipergpt,PICa,gupta2023visual} without task-specific model training or supervised finetuning of fundation models~\cite{PhilipIsola-FoundationBook, PhilipIsola-CVPR2024, PhilipIsola-ICML2024}.
%%\textbf{LLM Enhancement in Visual Reasoning.} Recent advancement in LLMs has fueled the generation of program scripts for visual inference
%%
The visual inference methods prompt an LLM to output an explicit sequence of operations that can invoke pre-built executable modules. ProgPrompt~\cite{10161317} uses an LLM to produce robot task plans as executable code given high-level instructions. 
PICa~\cite{PICa} introduces the representation of visual information as text via objects and their attributes detected, 
%%as well as image caption, 
and it improves the in-context learning with the additional textual data to GPT-3 to obtain answer to a visual question. 
ViperGPT~\cite{suris2023vipergpt} formulates visual questions as Python programs calling vision APIs to generate answers via code execution. VisProg~\cite{gupta2023visual} introduced a well-structured program instruction template for visual reasoning and an interpreter is 
to execute the external pre-trained vision moddel or a Python module in Python library. 
GENOME~\cite{chen2023genome} extends Visprog to unseen task scenarios by adding a process to create new modules and new in-context learning examples.  However, most existing zero-shot approaches suffer from the problem of blindly entrusting LLM generated programs, instead of integrating error checking and repairing with result verification as the preconditions for invoking program execution, minimizing the detrimental effect of logical errors in LLM-generated programs, such as incorrect planning steps, non-existent external modules due to undesirable LLM hallucination.

\end{document}